%% file: main.tex
\documentclass[11pt]{article}

\usepackage[preprint]{acl}

\makeatletter
\@ifpackagelater{lineno}{2026/02/11}{%
  \RemoveFromHook{build/column/before}[lineno]%
  \AtBeginDocument{\def\@LN@makecol{\@LN@orig@makecol
    \setbox\@outputbox\vbox{\boxmaxdepth\@maxdepth
      \protected@write\@auxout{}{\string\def\string\@LN@column{\if@firstcolumn1\else2\fi}}%
      \box\@outputbox}}}%
}{}
\makeatother
\usepackage{times}
\usepackage{latexsym}
\usepackage[T1]{fontenc}
\usepackage[utf8]{inputenc}
\usepackage{microtype}
\usepackage{graphicx}

\input{src/math_commands.tex}

\usepackage{booktabs}
\usepackage{float}
\usepackage{multirow}
\usepackage{colortbl}
\usepackage{array}
\usepackage{graphicx}
\usepackage{fvextra}   
\usepackage{tcolorbox}\tcbuselibrary{breakable}  
\newtcolorbox{tmplbox}[2][]{breakable, colback=black!3, colframe=black!88, coltitle=white,
  fonttitle=\bfseries\small, arc=1.6mm, boxrule=0.6pt, left=4pt, right=4pt, top=3pt,
  bottom=3pt, toptitle=2.5pt, bottomtitle=2.5pt, title={#2}, #1}
\newcommand{\tmplhead}[1]{{\setlength{\fboxsep}{0.8pt}\colorbox{chprosebg}{\textbf{#1}}}}
\newcommand{\tmplph}[1]{\textcolor{black!50}{#1}}
\definecolor{chkeybg}{HTML}{D6E6F4}\definecolor{chredbg}{HTML}{EAF1F8}\definecolor{chdropbg}{HTML}{E8E8E8}

\usepackage{tikz}
\definecolor{arrtemplatecolor}{HTML}{A32638}

\newcommand{\coremem}{\textbf{CoReMem}}  
\graphicspath{{figs/}}

\newcommand{\placefig}[3]{\IfFileExists{figs/#1}{\includegraphics[width=#2]{#1}}{\fbox{\parbox[c][#3][c]{\dimexpr#2-2\fboxsep-2\fboxrule\relax}{\centering\small\texttt{figs/\detokenize{#1}}\\ placeholder --- drop the PNG here}}}}
\usetikzlibrary{positioning}
\usepackage{amsmath,amssymb}
\makeatletter\renewcommand{\eqref}[1]{\textup{\tagform@{\ref{#1}}}}\makeatother
\usepackage{amsthm}
\newtheorem{proposition}{Proposition}
\theoremstyle{definition}
\newtheorem{definition}{Definition}
\usepackage{algorithm}
\usepackage{algorithmic}
\definecolor{chprose}{RGB}{192,132,37}\definecolor{chprosebg}{RGB}{251,236,210}
\definecolor{chcmd}{RGB}{58,124,176}\definecolor{chcmdbg}{RGB}{216,233,247}
\definecolor{chobs}{RGB}{122,90,166}\definecolor{chobsbg}{RGB}{236,228,245}
\definecolor{chrc}{RGB}{91,63,133}\definecolor{chrcbg}{RGB}{228,216,240}
\definecolor{chdiff}{RGB}{78,140,74}\definecolor{chdiffbg}{RGB}{221,238,220}
\definecolor{chfail}{RGB}{181,84,84}\definecolor{chfailbg}{RGB}{246,223,223}
\definecolor{memblue}{RGB}{219,238,254}
\definecolor{cbest}{RGB}{198,228,252}
\definecolor{csec}{RGB}{229,242,253}
\newcolumntype{C}{>{\centering\arraybackslash}m{4.8em}}  
\newcolumntype{M}{>{\centering\arraybackslash}m{5.0em}}  
\newcommand{\mmc}[2]{\begin{tabular}[c]{@{}c@{}}#1\\(#2)\end{tabular}}
\definecolor{gainblue}{RGB}{0,70,190}      
\definecolor{lossblue}{RGB}{150,190,235}   
\newcommand{\dg}[1]{{\tiny\textcolor{gainblue}{\textbf{#1}}}}   
\newcommand{\dl}[1]{{\tiny\textcolor{lossblue}{\textbf{#1}}}}
\newcommand{\dz}[1]{{\tiny\textcolor{gray}{\textbf{#1}}}}
\newcommand{\hang}[1]{\makebox[0pt][l]{\,\raisebox{0.55ex}{#1}}}
\newcommand{\sdsub}[1]{\raisebox{-0.45ex}{\tiny$\pm$#1}}   
\definecolor{cstale}{RGB}{140,140,140}

\definecolor{chestnut}{cmyk}{0, 0.7808, 0.4429, 0.1412}

\definecolor{phlcolor}{RGB}{128,40,150}
\definecolor{fieldkey}{HTML}{1F4E79}\definecolor{fieldred}{HTML}{2E6DA4}\definecolor{fielddrop}{HTML}{898781}

\usepackage{pifont}

\definecolor{bguse}{HTML}{F7EFD9}\definecolor{bgval}{HTML}{D6E6F4}
\newcommand{\hluse}[1]{{\setlength{\fboxsep}{1.5pt}\colorbox{bguse}{#1}}}
\newcommand{\hlval}[1]{{\setlength{\fboxsep}{1.5pt}\colorbox{bgval}{#1}}}
\newcommand{\mhluse}[1]{\hluse{$\displaystyle #1$}}
\newcommand{\mhlval}[1]{\hlval{$\displaystyle #1$}}
\definecolor{cmem}{HTML}{1F4E79}\definecolor{cses}{HTML}{8A6012}\definecolor{cfree}{HTML}{2F7D5B}\definecolor{cstep}{HTML}{6B3FA0}
\newcommand{\cstep}[1]{\textcolor{cstep}{#1}}
\definecolor{bggate}{HTML}{E9E1F3}
\newcommand{\mhlgate}[1]{{\setlength{\fboxsep}{1.5pt}\colorbox{bggate}{$\displaystyle #1$}}}
\newcommand{\hlgate}[1]{{\setlength{\fboxsep}{1.5pt}\colorbox{bggate}{#1}}}
\newcommand{\cmem}[1]{\textcolor{cmem}{#1}}
\newcommand{\cses}[1]{\textcolor{cses}{#1}}
\newcommand{\cfree}[1]{\textcolor{cfree}{#1}}
\definecolor{intrologiccolor}{HTML}{007C83}

\title{CoReMem: Cost-Internalizing Memory \\ for Successive Coding Tasks with Prospective Valuation}

\author{
 \textbf{Peiwen Li\textsuperscript{1}},
 \textbf{Harshit Verma\textsuperscript{1,*}},
 \textbf{Sophia Zhou\textsuperscript{1,*}},
 \textbf{Haiwen Wang\textsuperscript{1}},
 \textbf{Ngoc Bui\textsuperscript{1}},
\\
 \textbf{Heng Chang\textsuperscript{2}},
 \textbf{Fei Fei\textsuperscript{3}},
 \textbf{David van Dijk\textsuperscript{1}},
 \textbf{Rex Ying\textsuperscript{1}}
\\
 \textsuperscript{1}Yale University,
 \textsuperscript{2}Tsinghua University,
 \textsuperscript{3}Goodvision AI
\\
 \small{\textsuperscript{*}Equal contribution. \textbf{Correspondence:} \href{mailto:peiwen.li@yale.edu}{peiwen.li@yale.edu}}
}

\definecolor{citeblue}{HTML}{2E6DA4}
\hypersetup{citecolor=citeblue,urlcolor=citeblue,linkcolor=black}
\begin{document}

\maketitle

\begin{abstract}
Coding agents tackle successive tasks on an evolving repository, carrying forward edits and writing memory before the next task is known.
Selecting memory for unknown next tasks requires balancing future usefulness against the costs of writing and reading it.
The actor must also decide when acting on available memory saves effort as the task unfolds.
We introduce \underline{\textbf{Co}}ding-\underline{\textbf{Re}}pository \underline{\textbf{Mem}}ory \mbox{(\coremem{})}, a cost-internalizing framework that guides memory design and actor adaptation through measured memory value on the next task.
Prospective segment valuation compares runs of the next task with and without individual trajectory segments from the same inherited repository state, guiding a compact memory built around \emph{anchors}: repository locations and executable checks.
Actor adaptation estimates on-policy memory value within rollout groups with and without memory, using a reward that prioritizes resolution and penalizes step count in successful runs.
Provenance-gated value credit assigns the additional value term to steps identified as using memory-supplied anchors, while every step retains outcome credit.
On ChainSWE, \coremem{} raises the 27B actor's resolve rate from $21.0\%$ to $25.7\%$, with $17\%$ fewer steps and $11\%$ fewer total cache-aware tokens, including memory-module calls, relative to the untrained actor model without memory.
\end{abstract}

\section{Introduction}\label{sec:intro}
\begin{figure}[t]
\centering
\placefig{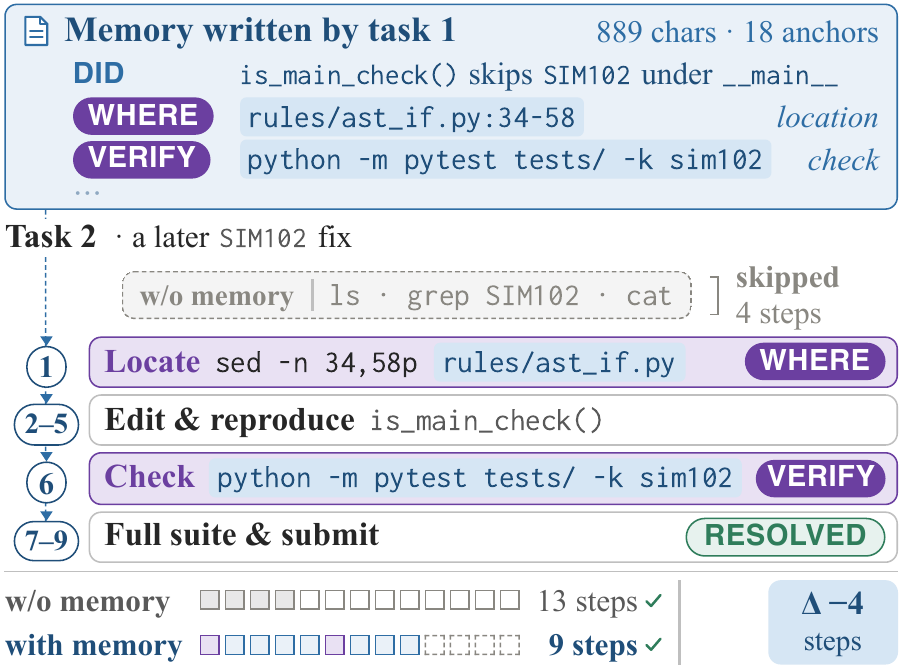}{0.85\columnwidth}{1.9in}
\caption{\textbf{Memory on the next task.} A memory written at the end of one flake8-simplify task
hands two anchors, a location (\texttt{WHERE}) and a check (\texttt{VERIFY}), to the next task on the
same repository; the post-trained actor model acts on them at steps 1 and 6 and skips the exploration it
would otherwise start with (step counts illustrative).}
\label{fig:case}
\end{figure}
\begin{figure*}[t]
\centering
\placefig{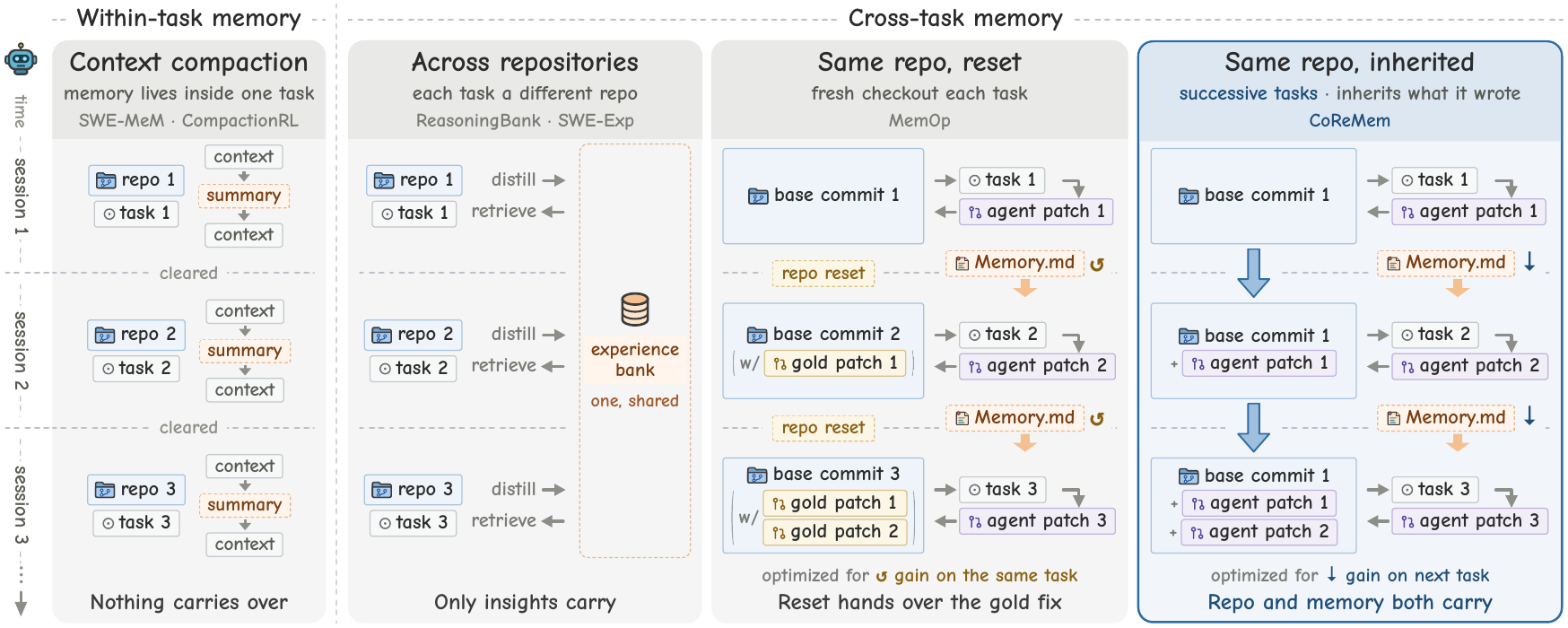}{0.95\textwidth}{2.0in}
\caption{\textbf{What crosses coding-task boundaries and where memory is valued and optimized.}
Existing settings keep memory within one task, transfer general experience across independent
tasks, or carry memory while resetting the repository. Our setting carries both repository state
and memory to unseen future tasks. The $\circlearrowleft$ and $\downarrow$ symbols indicate whether
memory is valued and optimized on its source task or on future tasks.}
\label{fig:setting}
\end{figure*}

In real-world software development, coding agents are expected to handle successive tasks on an
evolving repository. While prior research has primarily studied repository issues independently
\citep{swebench2024}, recent work considers sequences of tasks on a shared, evolving repository
\citep{chainswe2026}. We study this successive-task setting, in which tasks arrive sequentially,
each inherits the repository state left by earlier tasks, and future tasks are unknown in advance.
Since this state carries over, suitable memory can let agents reuse information acquired during
earlier exploration and avoid rediscovering it on unseen future tasks (Figure~\ref{fig:case}). We
therefore consider both
what a memory is worth and how an agent can use it to \emph{reduce the cost} of solving unseen future
tasks.

Existing memory methods for coding agents are not specifically designed for the above successive-task
setting and do not reliably save effort (steps and tokens) on unseen future tasks
(Figure~\ref{fig:prelim}). As illustrated in Figure~\ref{fig:setting}, within-task methods learn to
compact context for the current task, whose goal is already known
\citep{swemem2026,li2026compactionrl}. Most cross-task methods distill and transfer general
experience and insights across independent tasks and repositories
\citep{reasoningbank2025,sweexp2025}. MemOp instead accumulates memory across tasks within the same
repository, but since the repository state is reset between tasks in its setting, it optimizes its
memory design based on gains measured on the task that produced the memory \citep{memop2026}, rather
than on its value to an unseen future task that inherits the changed repository. Given these
mismatches in setting and objective, \textbf{Challenge~1} is to determine, before the future task is
known, \emph{what information is worth carrying across the task boundary}.

Memory's value is realized only when the actor model uses it appropriately during execution.
Existing coding memory methods mainly optimize what is stored or retrieved, while typically
leaving the model that reads it unchanged \citep{mem0_2025,reasoningbank2025,memop2026}. Yet an agent's
information needs evolve as a task unfolds \citep{repocoder2023,shen2026subtaskmem}, making
task-start retrieval insufficient to determine how memory should be used during execution. \textbf{Challenge~2} is therefore \emph{how to use memory efficiently during execution to reduce the
cost of solving unseen future tasks on the inherited repository}.

To address these challenges, we introduce \underline{\textbf{Co}}ding-\underline{\textbf{Re}}pository
\underline{\textbf{Mem}}ory (\coremem{}), a cost-internalizing memory framework for successive coding
tasks.
Specifically, prospective segment valuation of memory addresses Challenge~1 by first formulating
the \emph{prospective value} of a memory at a task boundary, then valuing each part of the current
task's trajectory by what it saves on the next task, and from these values inducing, through a
systematic procedure, a value-guided memory design for successive coding tasks. As a result, the
value concentrates in \emph{anchors}, references to repository locations and to executable checks,
which later also mark the steps that act on the memory. To address Challenge~2, we propose on-policy adaptation for memory use with provenance-gated
value credit, which adapts the actor model itself to act on the memory where that saves effort, in
two steps. Cost-internalizing on-policy valuation trains the actor model in groups of rollouts with
the memory, without it and with a misplaced memory, under a reward that charges steps inside success,
so that the average gap between the rollouts with the memory and those without it exposes the
memory's value to the current actor model. For the rollouts that run with the
memory, step-level provenance-gated value credit then pays the memory value to the steps whose anchors
came from the memory, so that memory use is pushed by its measured value while every step keeps its
outcome credit, and we show analytically that this push vanishes when the same value is paid to
every step.

Our contributions are summarized as follows.
\begingroup\setlength{\leftmargini}{1.3em}%
\begin{itemize}\itemsep0pt\topsep2pt
\item We propose prospective segment valuation of memory, a systematic procedure that values each
part of a trajectory by what it saves on the next task and induces a value-guided memory design
for successive coding tasks.
\item We propose on-policy adaptation for memory use with provenance-gated value credit, which
measures the memory value on-policy within each group of rollouts and passes the memory's value
directly to the steps that act on it.
\item On ChainSWE, a benchmark of successive tasks, \coremem{} raises resolve per token over no memory
by $41\%$ on average across three different actor models.\footnote{Code is available at an \href{https://anonymous.4open.science/r/CoReMem-F728/}{anonymous link}.} 
\end{itemize}
\endgroup

\section{Preliminaries}\label{sec:prelim}
\label{sec:setting}%
\begin{figure}[t]
\centering
\placefig{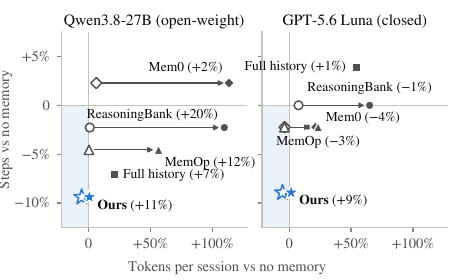}{\columnwidth}{1.55in}
\caption{\textbf{Steps and tokens of each memory form against no memory.} Hollow marks count only
the acting model's tokens, and the arrows add the tokens of the memory module.
Parentheses give the relative change in resolve rate. The star is \coremem{} w/o training (\S\ref{sec:replay}), the one form near the
shaded quadrant where steps and total tokens both fall.}
\label{fig:prelim}
\end{figure}

In this paper, we consider a repository worked on by a \emph{chain} of tasks $T_1,\dots,T_K$, carried
out one after another in a single persistent workspace~\citep{chainswe2026}. Each task runs as its own
session of an \emph{actor}, the model that acts in the repository, which leaves the repository state $S_k$ its edits produced and the trajectory
$\tau_k$ of its actions and observations. At the \emph{boundary} $b_k=(S_k,\tau_k,T_{k+1})$ the state
persists, since the next task opens the tree this one closed, while the trajectory is lost, and a
\emph{memory} $m=\pi_{\mathrm{write}}(\tau_k)$, written once at the end of task $k$ and before
$T_{k+1}$ is issued, is the only text that crosses. Unlike cross-episode memory that resets each task
to a base commit~\citep{memop2026, vibemembench2026}, the state here is \emph{inherited}
(Figure~\ref{fig:setting}), so a memory cannot be checked on its own task, whose answer the
repository holds.

\paragraph{Does existing memory save effort?}\label{sec:motivation}
We run the existing forms of cross-task memory, the full history of the earlier tasks
(App.~\ref{app:setup}), Mem0's facts~\citep{mem0_2025}, MemOp's guidebook~\citep{memop2026} and
ReasoningBank's strategies~\citep{reasoningbank2025}, on the ChainSWE test set and count, for every
form, the tokens of the acting model and of its memory module. Figure~\ref{fig:prelim} places each by its change in
steps and tokens against no memory, first with the acting model's tokens alone and then
adding the tokens of the memory module (Tables~\ref{tab:ledger} and~\ref{tab:closed} have every number).
Memory on successive tasks should lower the next task's effort, its steps and the tokens they cost, yet
no existing form reaches the quadrant where steps and total tokens both fall. Some buy resolution
rather than
effort, and each spends more tokens on its memory than it saves. Our aim is therefore a memory that lowers the effort of unseen future tasks without harming
resolution, with what it keeps derived by valuing each part of a finished session by the effort it
saves on these tasks and keeping what pays (\S\ref{sec:replay}, the star in Figure~\ref{fig:prelim}).

\paragraph{Learning objective.}\label{sec:formulation}
A memory can only point at what the next task should reuse, and effort is saved
only when the actor acts on it. An actor is trained to solve a task from scratch rather than to read
what an earlier task left, so it may pass over a memory that would spare it work or follow one that does
not apply~\citep{liu2024lost, shi2023distracted}, and whether
a memory applies becomes clear only as the actor works through the task. We therefore also adapt the actor's parameters from $\theta_0$ to $\theta=\theta_0+\Delta\theta$.
The next task's session is then $\tau\sim\pi_\theta(\cdot\mid S_k,T_{k+1},m)$. With the memory's
form fixed (\S\ref{sec:replay}), we learn $\Delta\theta$ so that on training boundaries
$\mathcal{B}$ the next task takes fewer steps $n(\tau)$ without resolving less often than the base actor,
with $\mathbb{E}_\theta$ over $b_k\in\mathcal{B}$ and $\tau\sim\pi_\theta$,
\begin{equation}\label{eq:objective}
\begin{aligned}
\min_{\Delta\theta}\;\;& \mathbb{E}_\theta\big[n(\tau)\big]\\
\text{s.t.}\;\;& \mathbb{E}_\theta\big[\mathrm{res}(\tau)\big]\geq\mathbb{E}_{\theta_0}\big[\mathrm{res}(\tau)\big],
\end{aligned}
\end{equation}
with $\mathrm{res}(\tau)\in\{0,1\}$ whether $\tau$ resolves $T_{k+1}$.

\section{Cost-Internalizing Memory}\label{sec:method}
\begin{figure*}[t]
\centering
\sbox0{\includegraphics[width=\textwidth]{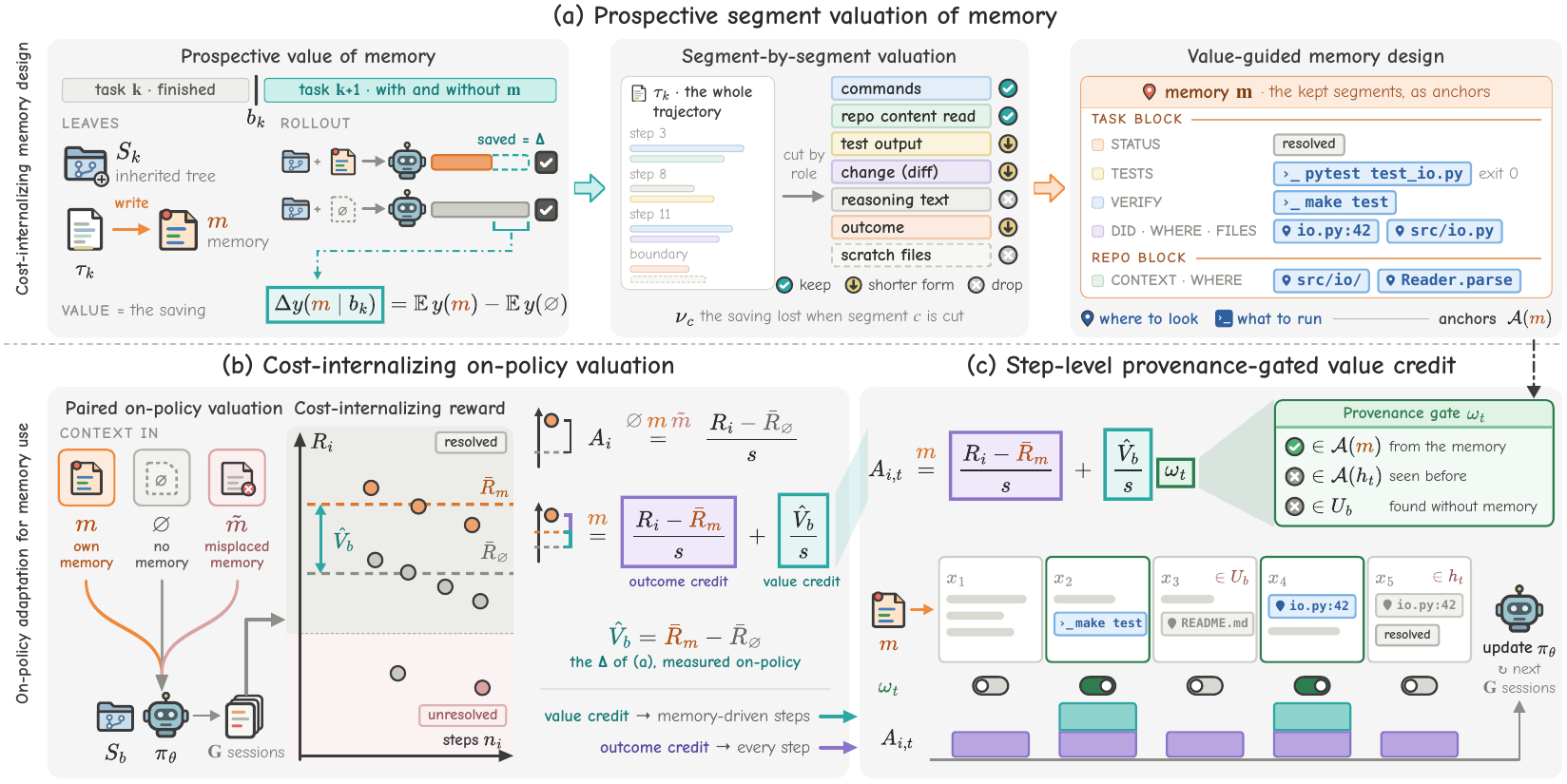}}\usebox0
\caption{\textbf{Overview of \coremem{}.} \textbf{(a)} Prospective segment valuation of memory. A
memory $m$ written from the trajectory $\tau_k$ is valued by the change $\Delta_y(m\mid b_k)$ it causes
on task $k{+}1$, run from the inherited tree $S_k$ with and without $m$. Removing each of the seven
segments of $\tau_k$ gives its value $\nu_c$, and the segments that pay are kept as anchors
$\mathcal{A}(m)$. \textbf{(b)} Cost-internalizing on-policy valuation. A group of $G$ sessions mixes
sessions with their own memory ($m$), no memory ($\varnothing$) and a misplaced memory ($\tilde m$)
under the reward $R_i$, and
$\widehat V_b=\bar R_m-\bar R_\varnothing$ is the memory's value on-policy. \textbf{(c)} Step-level
provenance-gated value credit. Only the steps that the gate $\omega_t$ marks as acting on an anchor
from the memory receive the value credit.}
\label{fig:design}
\end{figure*}
\coremem{} is cost-internalizing in the sense of Eq.~\eqref{eq:objective}. The effort a memory changes
on the next task is measured, and that measurement enters both what the memory keeps and how the
actor model is adapted to use it, in two parts (Figure~\ref{fig:design}).
\S\ref{sec:replay} defines the value of a memory on unseen future tasks and lets it guide what the
memory keeps, which turns out to be anchors. \S\ref{sec:grpo} adapts the actor model to use that memory
during execution, measuring the value again on-policy and paying it, through the anchors, to the steps
that use the memory, on top of the outcome credit.

\subsection{Prospective segment valuation of memory}\label{sec:replay}
To decide, before the next task is known, what to carry across a task boundary, we provide a
systematic procedure for inducing a memory design. We first define a memory's prospective value on
the next task, and use it to find which parts of a session's history are worth carrying because they
save effort on the next task.

\paragraph{Prospective value of memory.}\label{sec:apcv}
In real-world deployment, a memory is usually written for a task not yet issued, so we value it, and
train the actor to use it, on the unseen task that follows rather than on the task that produced it. Let
$y(m)$ be an outcome of the next task's session, which runs $T_{k+1}$ from the repository state $S_k$ that
task $k$ left, with memory $m$ in its context, namely whether it resolves, its steps or its actor tokens. The prospective value of a memory $m$
at boundary $b_k$ is the change it causes,
\begin{equation}\label{eq:price}
\Delta_y(m\mid b_k)=\mathbb{E}\big[y(m)\big]-\mathbb{E}\big[y(\varnothing)\big],
\end{equation}
where $m=\varnothing$ stands for running with no memory and the expectation is over the actor's
sampling. We estimate it as
$\widehat\Delta_y(m\mid b_k)=\bar y(m)-\bar y(\varnothing)$ from the mean outcomes of the runs, and the $n$
candidates at a boundary share one no-memory arm, $n{+}1$ arms instead of $2n$. We use this value both to attribute value across a session's
trajectory (below) and in on-policy actor training (\S\ref{sec:grpo}).

\begin{figure}[t]
\centering
\placefig{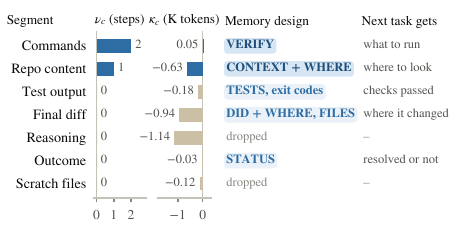}{\columnwidth}{1.5in}
\caption{\textbf{From segment values to the memory design.} The median change in steps ($\nu_c$) and
in thousand actor tokens ($\kappa_c$) when one segment is removed from the current task's trajectory
(Eq.~\eqref{eq:channel-value}; 27B actor model). \textcolor{fieldkey}{Blue} bars mark segments whose
removal adds steps. The right columns give the field each segment becomes in the memory design, kept as
a main field (\textcolor{fieldkey}{blue}), in reduced form (\textcolor{fieldred}{light blue}) or dropped
(\textcolor{fielddrop}{gray}), and what the next task gets from each of them.}
\label{fig:channels}
\end{figure}
\paragraph{Segment-by-segment valuation.}\label{sec:attrib}\label{sec:schema}\label{sec:writer}
We
measure which parts of a trajectory are worth carrying and let the measurements guide the form
(App.~\ref{app:derivation}).

\emph{(a) Cutting the trajectory into segments.} We split a trajectory into the seven segments of
Figure~\ref{fig:channels}, a segment being every part of the trajectory that plays one role, such as
the actor's reasoning text, its commands or the test output (Figure~\ref{fig:example-traj} in
App.~\ref{app:channels} shows a real excerpt labeled by segment). The split involves no prior knowledge about what matters, since the record of any coding agent
already comes in these parts, the actor's own text and commands, the observations the environment
returns, the diff it leaves and the outcome (each described in App.~\ref{app:channels}), and how much
of each is worth keeping is left to the valuation in (b) and (c).

\emph{(b) Valuing each segment.} Let $M=\tau_k$ be the current task's trajectory injected whole as a memory and
$M_c$ its segment $c$. With the prospective value of Eq.~\eqref{eq:price} at each boundary $b$, in steps ($\mathrm{s}$)
and in actor tokens ($\mathrm{t}$), removing $c$ from $M$ changes
\begin{equation}\label{eq:channel-value}
\begin{aligned}
\nu_c&=\operatorname{med}_b\big[\widehat\Delta_{\mathrm{s}}(M\!\setminus\! M_c)-\widehat\Delta_{\mathrm{s}}(M)\big],\\
\kappa_c&=\operatorname{med}_b\big[\widehat\Delta_{\mathrm{t}}(M\!\setminus\! M_c)-\widehat\Delta_{\mathrm{t}}(M)\big].
\end{aligned}
\end{equation}
In words, $\nu_c$ and $\kappa_c$ are the median changes in steps and in tokens when $c$ is removed, which shows
how much of the effect of $M$ on the next task, a saving or a cost, depends on each segment.

\emph{(c) From values to design.} We keep a segment if removing it adds steps on the next task, counting
$\nu_c$ as positive when it survives a permutation test corrected across segments, and drop one that
only adds tokens, unless a shorter form of it saves as many steps, in which case we keep that form.
Each $\nu_c$ is conditional on the other segments being present (Table~\ref{tab:derivation-channels}
gives each), so we then check the assembled memory against $M$ as a whole, on steps here and on resolution
in deployment (\S\ref{sec:exp}).

\paragraph{Value-guided memory design.} Removing the commands or the repository content read adds
steps on the next task, while removing the reasoning text or the diff body adds no steps and saves tokens
(Figure~\ref{fig:channels}). Shorter forms also work, as test commands with their exit codes save more steps than the full test
output and the diff's file list comes close to the full diff (App.~\ref{app:derivation}).
What saves the next task effort is
where to look and what to run, not the narrative, so we build the memory from the references that
carry it and define them below as anchors.

\begin{definition}[Anchor]\label{def:anchor}
An \emph{anchor} is a reference, taken from a trajectory, to a specific place in the repository, a
\emph{location}, or to a specific command that runs or tests the code, a \emph{check}, which the
next task can act on as it stands. $\mathcal{A}(x)$ denotes the anchors that a text $x$ contains.
\end{definition}
\noindent Locations are file paths with their lines and code names, such as
\texttt{rules/ast\_if.py:34-58}, and checks are commands that run or test the code and test
identifiers, such as \texttt{python -m pytest tests/ -k sim102} (App.~\ref{app:anchors}).

\noindent Following (c), the template keeps each segment in the form Figure~\ref{fig:channels} lists, the
commands and the repository content as main fields, the test output, the final diff and the outcome in
reduced form, and it drops the reasoning text and the scratch files (App.~\ref{app:templates}). Its main
components are the checks and the locations of Definition~\ref{def:anchor}, which potentially help the
next task save the effort of finding what to run and where to look, and they mirror the validation and
localization phases of Agentless~\citep{xia2024agentless}, which resolves issues cheaply without
open-ended exploration. We further organize these components into two blocks. The task block holds
what concerns the current task, what it changed and checked, and the repository block holds general
context, how the repository is laid out and tested. The memory keeps the whole median step saving of
$M$ with $6\%$ of its bytes. The procedure needs only paired runs on the next task and a split of the
trajectory by role, so it applies wherever tasks follow one another on shared state, and the segments
and anchors above are what it yields for coding. Anchors also make the memory traceable, since they
mark the steps of a later session that use it, which \S\ref{sec:grpo} builds on.

\subsection[On-policy adaptation for memory use with provenance-gated value credit]{On-policy adaptation for memory use\\with provenance-gated value credit}\label{sec:grpo}
To use memory efficiently during execution, at steps that are not known in advance but emerge as the
task unfolds (App.~\ref{app:prov-calib}), we adapt the actor model itself, so that it acts on the
memory in its context where that saves effort, as Eq.~\eqref{eq:objective} asks. To
separate what the memory contributes from how a session went, each training group mixes sessions
with and without the memory, so the memory's value to the current actor model is measured on-policy, and
that value is credited to the steps that acted on the memory.

\subsubsection{Cost-internalizing on-policy valuation}\label{sec:objective}
\paragraph{Cost-internalizing reward.} At a training boundary $b=b_k$ the current policy $\pi_\theta$ rolls out a group of $G$ sessions from the state
$S_b=S_k$, with and without memory. A session is scored on whether it resolves the task
and on how many steps it took,
\begin{multline}\label{eq:reward}
R_i \;=\;
\begin{cases}
W - n_i & \text{if session } i \text{ resolves } T_{k+1},\\[2pt]
\alpha\, f_i & \text{otherwise,}
\end{cases}
\\
W \;>\; n_{\max} + \alpha ,
\end{multline}
with $n_i$ its step count, $n_{\max}$ the step limit, $f_i\in[0,1]$ the fraction of the next task's
fail-to-pass tests that pass after the session, $\alpha$ the weight of that partial progress, and $W$ a payoff for
resolving that exceeds any step count plus $\alpha$. Every resolved session outranks every unresolved
one and effort is charged only inside success, so within a group no saving of steps outweighs a
resolution, which also removes the incentive to give up early. The reward thus works toward the constraint of
Eq.~\eqref{eq:objective}. Sessions with and
without memory are scored on this same scale, so a memory is charged for the steps it adds and
credited for the steps it saves and the resolutions it wins, which is what \emph{cost-internalizing}
means for the reward.

\paragraph{Paired on-policy valuation.}\label{sec:rollouts}\label{sec:dvr}
The group mixes three kinds of sessions, named by the memory they carry. Memory-free sessions
($\varnothing$) run with no memory, own-memory sessions ($m$) each carry a memory of $b$ drawn from the
memories $q_b$ the frozen base actor writes for it, and one misplaced-memory session ($\tilde m$)
carries a real memory written for another repository, right in form and wrong in content. A policy
that follows its memory blindly loses on $\tilde m$, and one that ignores memory forgoes what $m$ is
worth. Every session receives the ordinary group-relative advantage against the memory-free sessions
of its group,
\begin{equation}\label{eq:grpo}
A_i \;\overset{\varnothing,\,m,\,\tilde m}{=}\; \frac{R_i-\bar R_{\varnothing}}{s} \;\overset{m}{=}\; \mhluse{\frac{R_i-\bar R_{m}}{s}} \;+\; \mhlval{\frac{\widehat V_b}{s}},
\end{equation}
with $\bar R_{\varnothing}$ and $\bar R_m$ the mean rewards of the $\varnothing$ and $m$ sessions and $s$
a fixed scale (App.~\ref{app:train}). For $\varnothing$ and $\tilde m$ the ordinary advantage is the whole credit,
paid to every step, so acting on the misplaced memory is penalized through the outcome it leads to.
For $m$ the whole credit further splits into two parts. The first is the session's
\hluse{\emph{outcome credit}}, the advantage of the session against the sessions that carry a memory of $b$.
It reflects how the session itself dealt with the task and, as in ordinary GRPO, every step
receives it, which trains the session to resolve the task in fewer steps. The second is the memory's
\hlval{\emph{value credit}}, which carries the contribution of the memory itself, what a memory of $b$ is
worth to the actor, since
$\widehat V_b=\bar R_{m}-\bar R_{\varnothing}$ estimates $\mathbb{E}_{m\sim q_b}[\Delta_R(m\mid b)]$, the
prospective value of Eq.~\eqref{eq:price} with the reward as the outcome $y$ averaged over the
memories of $b$, measured on-policy at every visit at no extra run. The value credit goes to the few memory-driven steps, as set out next.

\begingroup
\makeatletter
\renewenvironment{table}{\@dblfloat{table}}{\end@dblfloat}
\makeatother
\input{tables/table1_main.tex}
\endgroup
\subsubsection{Step-level provenance-gated value credit}\label{sec:cca}
\paragraph{Provenance gate.}\label{sec:cca-weight}
To give the memory's contribution directly to the few steps that act on the memory, where it can
change what the actor does, we identify those steps by a per-step signal of provenance, whether what
a step acts on came from the memory. A \emph{step} $t$ is one assistant message $x_t$ and the output
it receives. A step of an own-memory session ($m$) is
\hlgate{\emph{memory-driven}, $\omega_t=1$}, when \cstep{$x_t$ contains an anchor} $a$
(Definition~\ref{def:anchor}) that \cmem{comes from the memory}, \cses{was not seen before in the
current session}, and \cfree{is not found without memory},
\begin{equation}\label{eq:prov}
\begin{split}
\mhlgate{\omega_t}=\mathbf{1}\big[&\exists\,a\in\cstep{\mathcal{A}(x_t)}\colon\cmem{a\in\mathcal{A}(m)},\\
&\cses{a\notin\mathcal{A}(h_t)},\,\cfree{a\notin\mathcal{U}_b}\big],
\end{split}
\end{equation}
with $h_t$ the task and every message and observation before step $t$, the injected memory excluded,
and $\mathcal{U}_b$ the anchors that the memory-free sessions ($\varnothing$) of the same group reach
within their first few steps (calibration of the gate in App.~\ref{app:prov-calib}). The gate marks traceable use of the
memory, not the step's own effect on the outcome, which stays with the outcome credit of every step.

\paragraph{Step-level value credit.}\label{sec:cca-prop}
With this gate, each step $t$ of an own-memory session $i$ is credited
\begin{equation}\label{eq:value-credit}
A_{i,t} \;=\; \mhluse{\frac{R_i-\bar R_{m}}{s}} \;+\; \mhlgate{\omega_t}\,\mhlval{\frac{\widehat V_b}{s}},
\end{equation}
and each step's tokens are trained with that step's credit (App.~\ref{app:train}). Setting every
$\omega_t=1$ recovers the paired whole-trajectory credit of Eq.~\eqref{eq:grpo}.

\begin{proposition}[Value credit as a push on memory use]\label{prop:route}
Let $J^{\mathrm{mem}}_b$ be the expected reward of a session with a memory $m$ of boundary $b$ and
$V_b$ the memory's value under the current actor. Under the population form of
Eq.~\eqref{eq:value-credit}, the expected policy gradient on such sessions is
\begin{equation}\label{eq:prop}
g_m=\mhluse{\tfrac{1}{s}\nabla_\theta J^{\mathrm{mem}}_b}+\mhlval{\tfrac{V_b}{s}}\textstyle\sum_t\mathbb{E}\big[\nabla_\theta\mhlgate{p_t}\big],
\end{equation}
with $p_t=P_\theta(\omega_t{=}1\mid h_t,m)$ the probability that step $t$ is memory-driven, and its
second term vanishes when every $\omega_t=1$.
\end{proposition}
The first term is the outcome learning of Eq.~\eqref{eq:reward}, and the second moves how often the
actor acts on the memory, up where the memory pays and down where it does not; App.~\ref{app:cca-proof}
derives this.

\section{Experiments}\label{sec:exp}
\subsection{Experimental setup}\label{sec:setup}
We run ChainSWE~\citep{chainswe2026} in mini-swe-agent, with its $100$ task chains split into a
training and a test half over disjoint repositories. We compare our memory with running without
memory, with the full history of earlier tasks and with three memory layers, Mem0~\citep{mem0_2025},
MemOp~\citep{memop2026} and ReasoningBank~\citep{reasoningbank2025}, and report the resolve rate,
steps per task, tokens per session and resolve rate per token. The detailed experimental setup,
including the ledger, is in App.~\ref{app:setup}, and results on SWE-bench Verified, external task
chains and OpenHands are in App.~\ref{app:external}.

\subsection{Main results}\label{sec:main-results}
\begin{figure}[t]
\centering
\includegraphics[width=\columnwidth]{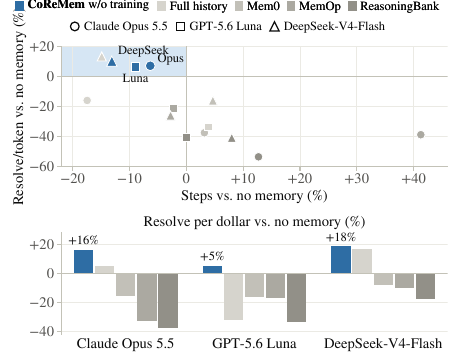}
\caption{\textbf{Steps, resolve per token and resolve per dollar against no memory for the API-served
models.} Top, each point is one memory layer on one model, and the shaded quadrant is where a memory
lowers steps and raises resolve per thousand total tokens, which \coremem{} w/o training reaches on all
three models. Bottom, the change in resolve per dollar at list prices, with full numbers in
Table~\ref{tab:closed} of App.~\ref{app:ledger-full}.}
\label{fig:layers}
\end{figure}

\paragraph{Our memory lowers effort while keeping resolution.}
With the actor model frozen, our memory lowers steps on all six actor models, the open-weight ones in
Table~\ref{tab:ledger} and the API-served ones in Figure~\ref{fig:layers}, significantly on three by
$9\%$ to $13\%$, without lowering resolution, which rises on five of them and is unchanged on Opus, and
at total tokens within $-7\%$ and $+5\%$ of running without memory on five of the six actor models
(tests in App.~\ref{app:setup}). Mem0, MemOp and ReasoningBank cost more tokens than our memory on
five of the six actor models they run on.

\paragraph{Post-training internalizes the cost.}
Post-training lowers steps against running without memory on all three actor models, by up to
$26\%$, at $9\%$ to $19\%$ fewer total tokens, and raises resolution by $22\%$ to $27\%$, so resolve
per total token rises by $38\%$ to $46\%$, $41\%$ on average (Table~\ref{tab:ledger}). Against the
same memory on the untrained actor model, steps fall by a further $24\%$ on 9B and $8\%$ on 27B.

\subsection{Detailed analysis}\label{sec:analysis}
We further conduct ablations on the three modules of \S\ref{sec:method}: what the memory layer (M1)
carries that the next task acts on, whether the cost-internalizing post-training (M2) lowers effort
beyond outcome learning, and whether the provenance-gated value credit (M3) changes how the actor
model treats memory; w/o training runs M1 on the frozen actor model,
resolve-only RL trains on the resolution term alone and M1$+$M2 trains without the value credit
(settings in App.~\ref{app:train} and~\ref{app:ablation}).
\begin{figure}[t]
\centering
\includegraphics[width=0.93\columnwidth]{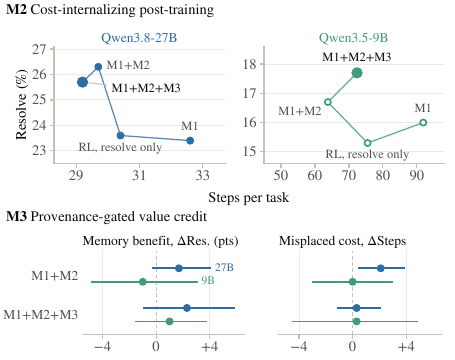}
\caption{\textbf{Evidence for M2 and M3.} Top, steps per task and resolve rate of the with-memory arms
in training order, where up and to the left is better and the larger marker is the full method.
Bottom, paired differences in the memory benefit to resolve rate and in the step cost of a misplaced
memory, blue for 27B and green for 9B as above, with $95\%$ intervals.}
\label{fig:modules}
\end{figure}
\paragraph{M1: the memory hands over anchors the next task uses.}
Our memory packs the most anchors per thousand injected characters, and the next task acts on $1.32$
of them per thousand characters against $0.92$ for MemOp, $0.19$ for ReasoningBank and $0.14$ for
Mem0 (Table~\ref{tab:anchors}).
\input{tables/table_anchor_use.tex}

\paragraph{M2: charging steps inside success lowers effort further.}
Figure~\ref{fig:modules} (top) places each variant on the effort--resolution plane. Resolve-only RL
lowers steps on 27B from $32.6$ to $30.4$ and on 9B from $91.9$ to $75.5$; charging steps inside
success (M1$+$M2) lowers them further, to $29.7$ and $63.8$, and the full method stays at $29.2$ and
$72.4$.

\paragraph{M3: the value credit teaches when to act on memory.}
Paying the memory's value to the steps that acted on it keeps the actor model from learning to follow
any memory it is given (Figure~\ref{fig:modules}, bottom): without the value credit a misplaced memory
costs the 27B actor model $2.1$ steps, while with it the cost shrinks to $+0.3$, and the memory benefit
grows from $+1.7$ to $+2.3$ points on 27B.

\section{Related Work}\label{sec:related}
\paragraph{Memory design and valuation.}
Coding memory spans cross-task experience \citep{reasoningbank2025}, within-task compression and trace validation \citep{swemem2026,memtrace2026}; MemOp evaluates memory through source-task replay and VibeMemBench through target-aware experience \citep{memop2026,vibemembench2026}.

\paragraph{Memory optimization.}
Reinforcement learning trains memory construction with outcome rewards \citep{memalpha2025,memoryr1_2025}, refines credit through shared-state rerollouts \citep{memoryr2_2026} and localizes it to selected turns or response tokens \citep{xie2026echo,memcalib2026}.
Our actor adaptation routes the memory's on-policy value to provenance-marked steps on top of the outcome credit that every step keeps (\S\ref{sec:grpo}).

\section{Conclusion}\label{sec:conclusion}
Memory for successive coding tasks is what one task leaves for a task it has not seen. Its worth,
measured by what the next task saves when it runs with the memory and without it, concentrates in a
few anchors, the locations and checks the next task would otherwise rediscover on its own. A memory
built around those anchors lowers the next task's effort on six actor models, and an actor model
trained on that value, paid where it acts on them, lowers effort further without lowering resolution.

\clearpage
\section*{Limitations}
\label{sec:limitations}
Our main experiments use Python repositories in the mini-swe-agent harness, and post-training covers
the three open-weight actor models. Other programming languages and larger actor models
remain to be tested. We measure effort by steps and cache-aware tokens and do not report wall-clock
latency or energy use. How memory should be accumulated and refreshed over longer repository
histories is future work.

\section*{Ethical Considerations}
\label{sec:ethics}
This work uses publicly released benchmarks, built from open-source repositories, and models in
accordance with their licenses, and involves no human participants. Every agent session runs inside
the per-task container image of its benchmark. In deployment, a cross-task memory may retain
repository content such as file paths, commands and test output, so in private repositories it should
follow the repository's access control and retention policy, with secrets redacted before storage. As
plain text, the memory can be audited and deleted.

\bibliography{ref_draft,ref_checked}

\clearpage
\appendix
\raggedbottom
\renewcommand{\dbltopfraction}{0.95}\renewcommand{\dblfloatpagefraction}{0.6}

\section{Training Details}\label{app:train}
Boundaries are drawn without replacement with probability proportional to $(u_b+\epsilon)\,g_b$.
Here $u_b$ is a running average of $\hat p_b(1-\hat p_b)$, where $\hat p_b$ is a running estimate of
how often the memory-free sessions at $b$ resolve, both at rate $\lambda{=}0.25$, $g_b$ drops to $0.25$
after a group at $b$ resolves nothing, and $\epsilon$ is a floor that keeps every boundary reachable. A
boundary the actor always or never resolves carries no outcome contrast, and the weight peaks where a
resolved and an unresolved session are equally likely. Boundaries that cannot be graded are removed
before training, and for the 27B actor also those that no session has ever resolved.
Each training group runs $G{=}8$ sessions at a boundary, four with no memory, three with memories of
$b$ written by the frozen base actor, drawn from a pool of three per boundary for the 27B actor and
twelve for the 9B, and one with a misplaced memory drawn from another repository.
The reward of Eq.~\eqref{eq:reward} uses $W{=}250$, with $n_{\max}{=}100$ and $\alpha{=}50$ for the 27B
actor and $n_{\max}{=}200$ and $\alpha{=}16$ for the 9B, and advantages
use the fixed scale $s{=}100$ rather than the group's standard deviation, which would inflate the step
differences of a group that resolves everything and would give the value credit a unit that changes
from group to group, whereas a fixed scale keeps $\widehat V_b/s$ comparable across boundaries.
Dropping the standard deviation follows Dr.~GRPO~\citep{liu2025drgrpo}.
The grouped return baseline is related to GRPO~\citep{shao2024deepseekmath} and
REINFORCE-style updates~\citep{ahmadian2024rloo}; our groups additionally separate memory conditions.
The policy follows the
advantages by policy gradient
with a KL penalty to the base actor.
The set $\mathcal{U}_b$ of Eq.~\eqref{eq:prov} is collected from the first
$K{=}8$ steps of the memory-free sessions. The actor is a LoRA adapter on the base actor, and with
$x_j$ an assistant token, $\mathrm{prefix}_j$ the realised history before it and $N$ the number of
assistant tokens in the update, the loss of session $i$ is
\begin{equation}\label{eq:route-rollout}
\ell_i(\theta) = -\tfrac{1}{N}\sum_{t\in\tau_i}A_{i,t}
\sum_{j\in t}\log\pi_\theta\!\left(x_j\mid\mathrm{prefix}_j,c_i\right),
\end{equation}
with $c_i\in\{\varnothing,m,\tilde m\}$ the memory session $i$ ran with ($\tilde m$ the misplaced one),
$A_{i,t}=A_i$ for sessions without their own memory, plus a KL penalty to the base actor with weight
$\beta{=}0.2$, summed over the same tokens and divided by $N$. For the 9B actor the policy term of a
session is instead divided by its own assistant tokens and by the number of sessions in the update.
Nothing is resampled for the per-step credit, and one forward and one backward pass serve both credits.
The adapter has LoRA rank and alpha $16$ and is trained with AdamW at a learning rate of
$2{\times}10^{-4}$ for the 27B actor and $10^{-4}$ for the 9B, with $3$ warmup updates, gradient
clipping at $1.0$ and sampling temperature $1.0$, for $20$ updates. Training sequences are truncated at
$32$K tokens for the 27B actor and $49$K for the 9B, sessions that do not return within $1{,}500$\,s
and $3{,}600$\,s respectively are left out of their group, and for the 9B so are sessions that end in a
repeated format error.

\begin{algorithm*}[t]
\caption{One update of on-policy adaptation for memory use with provenance-gated value credit.}
\label{alg:c3}
\begin{algorithmic}[1]
\REQUIRE actor $\pi_\theta$ as a LoRA adapter over the frozen base $\pi_{\mathrm{base}}$; training
boundaries $\mathcal{B}$ with estimates $\hat p_b$, $u_b$ and gates $g_b$; a pool of memories per boundary
written by $\pi_{\mathrm{base}}$ (three for 27B, twelve for 9B); a cross-repository donor pool
\STATE draw $8$ boundaries without replacement with probability proportional to $(u_b+\epsilon)\,g_b$
\FOR{each drawn boundary $b$}
  \STATE restore $S_b$ and run $G{=}8$ sessions of $\pi_\theta$ on $T_{k+1}$, four with no memory,
         three with memories of $b$, one with a misplaced memory from another repository
  \STATE grade every session and score it by Eq.~\eqref{eq:reward}
  \IF{every session of the group carries a grading defect}
    \STATE drop the group from the update and from $\hat p_b$ and $u_b$, and lower $g_b$ as for a group
           that resolves nothing
  \ELSE
    \STATE $\widehat V_b \leftarrow \bar R_{m}-\bar R_{\varnothing}$, and update $\hat p_b$ from the
           memory-free sessions, then $u_b$ and $g_b$
    \STATE $A_i \leftarrow (R_i-\bar R_{\varnothing})/s$ on every step of the memory-free and the
           misplaced-memory sessions, as in Eq.~\eqref{eq:grpo}
    \STATE collect $\mathcal{U}_b$ from the first $K$ steps of the memory-free sessions
           and mark $\omega_t$ on every step of the own-memory sessions by Eq.~\eqref{eq:prov}
    \STATE $A_{i,t} \leftarrow (R_i-\bar R_{m})/s+\omega_t\,\widehat V_b/s$ on the own-memory sessions,
           as in Eq.~\eqref{eq:value-credit}
  \ENDIF
\ENDFOR
\STATE update $\theta$ once on $\sum_i\ell_i$ of Eq.~\eqref{eq:route-rollout} plus the KL term, over the
       retained sessions with nonzero credit, taken in descending mean $|A_{i,t}|$ and skipping any that
       no longer fits, until the update's token budget ($350$K tokens for 27B, $600$K for 9B) is spent
\end{algorithmic}
\end{algorithm*}

\input{appendix_drafts/app_train_dynamics}

\section{Experimental Details}\label{app:setup}
\paragraph{Data.}
ChainSWE~\citep{chainswe2026} provides $100$ chains of successive tasks, each chain a sequence of real
issues of one repository solved in order on the tree the earlier tasks left. We split the chains by
repository, without looking at any outcome, into a training half of $52$ chains ($97$
boundaries) and a test half of $48$ chains ($96$ boundaries), so no repository appears in both. A test
boundary is linked when the files that the gold and test patches of $T_{k+1}$ touch overlap those
changed by earlier gold patches of its chain ($54$ boundaries) and unlinked otherwise ($42$). Training
and checkpoint selection use the training half and every comparison uses the test half.
The external
sets are described in App.~\ref{app:external}.

\paragraph{Actor models and harness.}
Six actor models run in mini-swe-agent with a $98$K-token window, three open-weight models that we also
post-train, Qwen3.5-9B~\citep{qwen35_2026}, Qwen3.8-27B-FP8~\citep{qwen38_2026} and
Gemma-4-31B~\citep{gemma4_2026}, with a step limit of $100$ for the 27B model
and $200$ otherwise, and three API-served models, Claude Opus 5.5~\citep{opus55_2026},
GPT-5.6 Luna~\citep{gpt56_2026} and DeepSeek-V4-Flash~\citep{deepseekv4_2026},
with a step limit of $200$.

\paragraph{Memory layers compared.}
No memory runs every next task from the inherited tree with its task description alone. Full
history injects the most recent part of the transcript of the earlier tasks.
Mem0~\citep{mem0_2025}, MemOp~\citep{memop2026} and
ReasoningBank~\citep{reasoningbank2025} run with their official instructions and their own write and
read calls, which the ledger counts on the memory side. \coremem{} w/o training is our memory, written
by the frozen actor model at the end of each current task's session (App.~\ref{app:templates}) and
injected into the next task's session, and \coremem{} is the post-trained actor model with the same memory
written by itself.

\paragraph{Metrics.}
Resolve rate is the share of next tasks whose hidden tests pass on the working tree at the end of
the session, reported overall and on the linked and unlinked boundaries. Steps is the number of
assistant turns of a session, each one command. Tokens per session count every call of the session,
cache-aware, the actor model's own calls (Actor) and the calls of the memory module (Writer), with
Total their sum. Resolve per token divides the overall
resolve rate by the thousand actor or total tokens per session. For the API-served models, USD per
session prices the same calls at the provider's list rates, DeepSeek-V4-Flash at DeepSeek's off-peak
rate, a deployment-free conversion and not an expense, and resolve per USD divides the resolve rate
by it.

\paragraph{Replays and tests.}
Every row is read on the mean over its replays, $12$ for the frozen 9B rows, $6$ to $7$
for 27B, $6$ for DeepSeek, $3$ for Gemma and Luna and $1$ for Opus. Differences against the no-memory row of the
same actor model are tested with paired chain-cluster bootstrap intervals over the $48$ test chains,
and the main text calls a difference significant when the $95\%$ interval excludes zero.

\paragraph{Post-training and checkpoint rule.}
The 27B and 9B actor models are post-trained with LoRA at a checkpoint chosen among the few saved
during a run, on the training split and by a rule fixed in advance, resolution no lower than the base
actor model's and then the fewest steps.
The Gemma actor model is post-trained for $8$ updates with no checkpoint selection.
App.~\ref{app:train} reports the 27B actor model at every checkpoint on the test sessions.

\section{Ablation Details}\label{app:ablation}
The ablation of \S\ref{sec:analysis} and Figure~\ref{fig:modules} reads each variant with its own
memory, without memory and with a misplaced memory, averaging $18/18/6$ replays of these arms for w/o
training, $18/9/6$ for M1$+$M2 and $6/6/3$ for the full method on 27B, and $3/3/3$ on 9B except the
full method with $6/3/3$; the 27B resolve-only arm comes from the run of App.~\ref{app:train} ($3$
replays in a later window). \emph{Memory benefit} pairs a variant's with-memory arm against its own
no-memory arm, and \emph{misplaced cost} pairs the same actor given another repository's memory against
that no-memory arm. Intervals are paired chain-cluster bootstrap $95\%$ intervals over the $48$ test
chains ($2000$ resamples, seed $0$), and an interval that excludes zero is what the main text calls
significant. On the frozen 27B actor model the memory raises resolution from $21.3\%$ to $23.4\%$
($+2.1$ $[+0.4, +3.7]$) and lowers steps from $35.3$ to $32.6$, and a misplaced memory changes steps
by $-0.1$ $[-1.7, +1.5]$. Without the value credit (M1$+$M2) a misplaced memory costs the 27B
actor model $+2.1$ steps $[+0.4, +3.9]$ and with it $+0.3$ $[-1.2, +2.1]$.
On 9B the with-memory and no-memory arms of the untrained, M1$+$M2 and
full variants were read in one window, their misplaced-memory arms and the resolve-only variant in later
ones. The trained variants differ from each other only in the reward and the credit rule; all share
learning rate, update count, group shape, memory pool and serving path, and every variant's untrained
control shares that serving path.

\section{Ledger of the API-served Models}\label{app:ledger-full}
Table~\ref{tab:closed} gives the ledger of the API-served models with the list-price cost per session;
the open-weight models are in Table~\ref{tab:ledger} and Figure~\ref{fig:layers} summarizes the API-served ones.
\begingroup
\makeatletter
\renewenvironment{table}{\@dblfloat{table}}{\end@dblfloat}
\makeatother
\input{tables/table_closed.tex}

\endgroup

\section{Results Beyond the Training Distribution}\label{app:external}\label{app:pending}
On SWE-bench Verified~\citep{openai2024swebenchverified}, one task per instance and no memory, the post-trained actor model resolves
$4.5$ points more than its untrained counterpart on 27B and $6.4$ on 9B, with $7\%$ and $22\%$ fewer
steps and $9\%$ and $22\%$ fewer actor tokens (Table~\ref{tab:ext-verified}). On two external task
chains run with the frozen 27B actor model and the same memory-writing instruction as on ChainSWE, our
memory lowers steps by $29\%$ on SWEContextBench~\citep{swecontextbench2026} and $14\%$ on SWE-Bench-CL~\citep{joshi2025swebenchcl}. On SWEContextBench, where
the next task inherits the current task's tree, it also lowers total tokens by $38\%$, and the
tightened instruction raises resolution by $8$ points over no memory; on SWE-Bench-CL,
where the memory alone crosses the boundary, resolution rises by $5$ points (Table~\ref{tab:ext-memory}).
Under the OpenHands harness~\citep{wang2025openhands}, with the same frozen 27B actor model and memory, our memory raises
resolution by $6$ points and is the one memory layer that raises resolve per token, by $11\%$,
and the post-trained actor model raises resolution by a further $5$ points, to $21.3\%$ (Table~\ref{tab:openhands}).

\begingroup
\makeatletter
\renewenvironment{table}{\@dblfloat{table}}{\end@dblfloat}
\makeatother
\input{tables/table_verified.tex}

\endgroup
\paragraph{SWE-bench Verified protocol.} Standard SWE-bench Verified harness images and test commands; resolve is the official FAIL\_TO\_PASS and PASS\_TO\_PASS
check. No memory in either row, since a Verified instance has no earlier task. The untrained row is the zero-update
adapter served on the same LoRA path as the trained row, not the native base, so that the serving path does not enter the difference.
Resolve grades the working tree at session end; grading only the submitted patch gives $68.9$ (untrained) and $73.3$ (post-trained) for 27B. The denominator is all $500$ instances; on the $498$ whose gold patch passes the dataset's own test command the differences are unchanged. Paired bootstrap over instances ($10{,}000$ resamples), 27B post-trained minus untrained: resolve $+4.5$ [$+2.4$, $+6.5$], steps $-2.2$ [$-3.1$, $-1.2$], actor tokens $-2.2$K [$-3.2$, $-1.3$], limit hits $-4.5$ pp [$-5.8$, $-3.2$]; ** marks a $99\%$ interval that excludes zero. Res/Tok is resolve (\%) per thousand actor tokens. One Verified instance (\texttt{pydata\_\_xarray-3305}) also appears in a ChainSWE test chain and none in a training chain. For 9B (post-trained: the full method at the pre-registered checkpoint, after $12$ updates), post-trained minus untrained: resolve $+6.4$ [$+4.0$, $+8.8$], steps $-20.2$ [$-22.4$, $-18.0$], actor tokens $-20.6$K [$-23.0$, $-18.1$], limit hits $-6.9$ pp [$-8.4$, $-5.6$], and context-window exits fall from $21.7\%$ to $5.8\%$; the gold-validated subset gives the same differences.

\paragraph{External task chains and harnesses.} Table~\ref{tab:ext-memory} asks whether a memory the
actor model writes for itself still lowers the next task's effort on successive tasks we did not mine, on
SWEContextBench with the inherited tree and on SWE-Bench-CL with the tree reset. Table~\ref{tab:openhands}
runs $94$ of the $96$ next tasks of the test split, the frozen and the post-trained 27B actor model and the same memory through OpenHands, with the harness alone changed.
\begingroup
\makeatletter
\renewenvironment{table}{\@dblfloat{table}}{\end@dblfloat}
\makeatother
\input{tables/placeholder_external_memory.tex}
\endgroup
\input{tables/table_openhands.tex}

\section{Proof of Proposition~\ref{prop:route}}\label{app:cca-proof}
\paragraph{Full statement.}
Fix a boundary $b$ with memories $q_b$ and veto set $\mathcal{U}_b$, let
$J^{\mathrm{mem}}_b(\theta)=\mathbb{E}_{q_b}\mathbb{E}_m[R]$ be the expected reward of a session that
runs with a memory of $b$ and $V_b=J^{\mathrm{mem}}_b-\mathbb{E}_{\varnothing}[R]$ the memory's value
under the current actor. Under the credit $A_t=(R-J^{\mathrm{mem}}_b)/s+\omega_t\,V_b/s$, the
population form of Eq.~\eqref{eq:value-credit}, the expected policy-gradient update on sessions that
carry a memory of $b$ reads
\[
\begin{split}
g_m \;&=\; \frac{1}{s}\,\nabla_\theta J^{\mathrm{mem}}_b(\theta)\\
&\;+\;\frac{V_b}{s}\sum_{t}\mathbb{E}_{q_b}\mathbb{E}_m\!\Big[\nabla_\theta P_\theta(\omega_t=1\mid h_t,m)\Big],
\end{split}
\]
where the gradient in the second term acts on the policy at step $t$ with the history $h_t$ held
fixed. Because $\omega_t$ depends on the step and its history and not on $\theta$, the expected score
of the steps it selects is the gradient of the probability of selecting them, so the memory's value
pushes how often the actor acts on the memory. Paid to every step instead, the same value multiplies
$\mathbb{E}[\nabla_\theta\log\pi_\theta(\tau\mid m)]=0$ and teaches nothing about memory use.

\paragraph{Setup.}
Fix the current parameters $\theta$ and condition on one memory $m\sim q_b$;
the statement follows by averaging over $m$, whose distribution does not depend on $\theta$. A session
$\tau\sim\pi_\theta(\cdot\mid S_b,m)$ is a finite sequence of steps, at each of which the actor emits a message $x_t$
after a history $h_t$, and the environment's responses do not depend on $\theta$. The provenance
$\omega_t=\omega(x_t,h_t)\in\{0,1\}$ of Eq.~\eqref{eq:prov} is a deterministic function of the
message, its history and the veto set, and does not depend on $\theta$. Write
$J_m=\mathbb{E}_m[R]$ for the expected reward under this $m$, take $J^{\mathrm{mem}}_b$ and $V_b$ as
in the proposition, constants that do not depend on the sampled $m$, and hold the veto set fixed.

\begin{proof}[Proof of Proposition~\ref{prop:route}]
The per-step credit defines the ascent direction
$g_m=\mathbb{E}_m\big[\sum_t A_t\nabla_\theta\log\pi_\theta(x_t\mid h_t,m)\big]$ with
$A_t=(R-J^{\mathrm{mem}}_b)/s+\omega_t V_b/s$, which the loss of Eq.~\eqref{eq:route-rollout} follows
up to the normalisation discussed below. For the first part, the environment does not depend on $\theta$, so
$\sum_t\nabla_\theta\log\pi_\theta(x_t\mid h_t,m)$ is the score of the session, its mean is zero, and
$\mathbb{E}_m\big[(R-J^{\mathrm{mem}}_b)\sum_t\nabla_\theta\log\pi_\theta(x_t\mid h_t,m)\big]=\nabla_\theta J_m$,
and averaging over $m\sim q_b$ gives $\nabla_\theta J^{\mathrm{mem}}_b$.
For the second, conditioning on $h_t$ and using that $\omega$ does not depend on $\theta$,
\[
\begin{aligned}
&\mathbb{E}\big[\omega(x_t,h_t)\,\nabla_\theta\log\pi_\theta(x_t\mid h_t,m)\,\big|\,h_t\big]\\
&\quad=\sum_{x}\omega(x,h_t)\,\nabla_\theta\pi_\theta(x\mid h_t,m)\\
&\quad=\nabla_\theta P_\theta(\omega_t=1\mid h_t,m),
\end{aligned}
\]
and taking the expectation over $h_t$, summing over $t$ and averaging over $m\sim q_b$ yields the
second term, with the constant $V_b$ passing through the average. Under
whole-trajectory credit $\omega_t$ is replaced by one, the same conditioning gives
$\sum_x\nabla_\theta\pi_\theta(x\mid h_t,m)=\nabla_\theta 1=0$ for every $t$, and the memory's value
contributes nothing.
\end{proof}

\paragraph{Remark on the cancellation.}
At every step
$\nabla_\theta P_\theta(\omega_t=1\mid h_t,m)+\nabla_\theta P_\theta(\omega_t=0\mid h_t,m)=0$. Paying
the memory's value to every step therefore adds, to the push toward the memory-driven actions, an equal
and opposite push toward all the others, and the two cancel in expectation while both remain in the
variance. The gated credit pays the first and withdraws the second.

\paragraph{Remark on the sample baselines.}
In training the two expectations are group means. On a memory-driven step the credit
is $(R_i-\bar R_{\varnothing})/s$, whose baseline is computed from other sessions and is independent
of the one it is subtracted from. On every other step the baseline $\bar R_m$ includes the session
itself, which scales the expected contribution of those steps by $1-1/n_m$, with $n_m{=}3$ the
number of own-memory sessions in the group, and leaves its direction unchanged. Under whole-trajectory
credit the same self-inclusion turns the value term into $\tfrac{1}{n_m s}\nabla_\theta J_m$, a rescaling of
the outcome gradient with no push specific to memory-driven steps. The loss of
Eq.~\eqref{eq:route-rollout} sums each step's credit over its tokens and divides by the update's token
count, so for the 27B actor its gradient is the per-step estimator of the proposition scaled by a
positive constant.
The own-memory sessions of a group carry different memories of $b$, so
$\widehat V_b$ estimates the value of a memory for $b$ written by the base actor, averaged over those
memories, which the proposition states with $q_b$.

\section{Provenance Calibration}\label{app:prov-calib}
The gate of Eq.~\eqref{eq:prov} reads the whole assistant message of a step, its reasoning text and its
command, so an anchor named in the reasoning text counts as much as one in the command, and each
anchor is credited at its first use. We calibrate provenance as training uses it, with the sibling veto over the first $8$ steps, on the
post-trained 27B actor's held-out sessions with its own memories ($288$ sessions over three replays).
Under a real memory $3.46\%$ of steps carry provenance and $169$ of the $288$ sessions have at least
one. Memory-free sessions scored against the memory they would have seen give a leakage null of
$1.61\%$ of steps, and misplaced memories give $1.05\%$.
Credited steps concentrate late in a trajectory, with deciles $7$ to $9$ at about
twice the rate of deciles $4$ to $6$ (ratio $2.02$ $[1.52, 2.78]$), where the actor runs tests and the
anchors it uses come from the memory's \texttt{TESTS} and \texttt{VERIFY} lines. Steps that match the
memory early, when the actor opens the files it names, are mostly steps that memory-free sessions also
take, so the veto removes about $88\%$ of them and leaves about $3\%$ of credited steps in the first
decile.
\paragraph{The early-window veto.}
The window $K$ was fixed before the arm was trained, on the $480$ own-memory rollouts of an earlier
training run. With no veto $5.5\%$ of steps are memory-driven, with the veto over the
first $8$ steps of the memory-free sessions $1.9\%$, and with a veto over their whole
length $0.65\%$. Memory-free sessions have a median of $26$ steps, so the first $8$ are an
early prefix, and a veto over their whole length would remove most of what the memory supplied.
In practice, a test command the memory lists is memory-driven at the step that first runs it, and not if the session ran it
earlier or the memory-free sessions reach it in their first steps; a memory-free session has
$\omega\equiv 0$.

\paragraph{Why not a likelihood ratio.}
The natural alternative weights a step by the ratio of its message's likelihood with and without the
memory in context. This alternative is related to context attribution~\citep{cohenwang2024contextcite}
and to probability contrasts used in context-aware decoding~\citep{shi2024cad} and
PMI-based summarization~\citep{vanderpoel2022pmi}; their targets are attribution or faithfulness rather
than executable-task effort. On this actor that ratio carries no signal. In a pilot of $36$ sessions on the 27B actor its
AUC is $0.506$ for real against the null, and the weight moves with message length rather than with
message content. A flat ratio next to a sharp anchor match says that the memory changes a few actions
rather than conditioning the whole distribution, which is why the credit signal is provenance and not
likelihood.

\section{Segments of a Trajectory}\label{app:channels}
Each step of a trajectory is an assistant message, made of reasoning text and a shell command, followed by the
output of that command. The command goes to the commands segment together with the exit code of its
step, and the output goes to one segment by what produced it, under rules fixed before the study (the
first match wins). Output of a command that prints a diff (\texttt{git diff}, \texttt{git show}, a
patch file) goes to the change segment; output of a test runner (\texttt{pytest}, \texttt{tox},
\texttt{unittest} and the like), of a script written in the session or of inline Python that contains an
\texttt{assert} goes to the test segment; all other output goes to the repository segment, including
that of linters such as \texttt{flake8}, which the list does not name. The change the task left in the tree joins
the change segment. The outcome and the scratch files come from the boundary, not from the transcript,
and harness format messages ($0.05\%$ of the material) are left out. Each segment is cut to a fixed
quota ($4{,}000$ characters, $200$ for the outcome and $800$ for scratch files), so a segment is the same
text in every arm that contains it. None of these assignments reads what the text says, since each
segment is read off what the harness records, the assistant message, the command and its exit code,
its output, the diff and the end of the session. Table~\ref{tab:channels-def} defines the seven segments
with an example from one real session, shown in Figure~\ref{fig:example-traj}.

\begin{table*}[t]
\centering\small
\setlength{\tabcolsep}{4pt}
\begin{tabular}{@{}>{\raggedright\arraybackslash}p{0.13\textwidth}>{\raggedright\arraybackslash}p{0.36\textwidth}>{\raggedright\arraybackslash}p{0.45\textwidth}@{}}
\toprule
\textbf{Segment} & \textbf{What it holds} & \textbf{In the example session (Figure~\ref{fig:example-traj})} \\
\midrule
Reasoning text & The actor's own text in each step, outside the command & ``Now let me verify the fix works with a test script.'' (step 8) \\
Commands & Every command the actor ran, in order, with the shell exit code of its step ($0$ for success, nonzero for failure) & \texttt{python -m pytest tests/ -k "sim102" -q} [exit 0] (step 9); a \texttt{grep} on a file that does not exist [exit 2] (step 2) \\
Repository content & Output of commands that read or search the repository, and of any other command that is not a test & The file list of \texttt{flake8\_simplify/rules/} and the two lines of \texttt{ast\_if.py} that mention \texttt{SIM102} (step 3); the \texttt{flake8} warning on the negative case (step 8) \\
Test output & Output of test runners, of scripts written in the session and of inline Python with an \texttt{assert} & \texttt{4 passed, 88 deselected} (step 9); \texttt{90 passed, 2 xfailed} (step 10) \\
Change & The diff the task left in the tree, and output of commands that print a diff & $23$ added lines in \texttt{ast\_if.py} that define \texttt{is\_main\_check} and call it (step 11) \\
Outcome & How the session ended, its exit status and step count & Submitted after $12$ steps \\
Scratch files & Paths the session's commands wrote, by shell redirection or from Python, and whether each still exists when the next task starts & \texttt{/tmp/test\_sim102.py} and \texttt{/tmp/test\_sim102b.py} (gone), \texttt{patch.txt} at the repository root (kept) \\
\bottomrule
\end{tabular}
\caption{\textbf{The seven segments of a trajectory.} The first five partition the transcript, and the
last two come from the boundary.}
\label{tab:channels-def}
\end{table*}

Three compressed forms are valued alongside the segments and decide how a kept segment is written: the
failing commands with their output (here steps 1 and 2, a search that found nothing and a \texttt{grep}
on a missing file), the tests with their exit codes but no output, and the file list of the change. The
exit code is recorded once per step, so when a command pipes into \texttt{tail} (steps 9 and 10) it is
the code of the pipeline, and the test's own result is in its output.

The example comes from the first task of a three-task chain on \texttt{flake8-simplify} in the training
split, run by the frozen 27B actor. The memory it wrote in its last turn is the one in
Figure~\ref{fig:example-note}, and its lines trace back to the trajectory as follows.
\texttt{FILES} and the \texttt{WHERE} of \texttt{DID} come from the change, whose added lines sit
exactly at lines $34$--$58$ of \texttt{ast\_if.py}. \texttt{STATUS} comes from the outcome and
\texttt{TESTS} from the full test run of step 10. The two \texttt{VERIFY} lines are the commands of
steps 9 and 8 with their exit codes. The \texttt{CONTEXT} lines come from the repository content read in
step 3 (one rule file per AST node type, \texttt{SIM102} in \texttt{ast\_if.py}) and in steps 4 to 6
(\texttt{constants.py}, \texttt{utils.py}), and from the test output of step 10 (\texttt{90 passed, 2
xfailed}). None of the reasoning text reaches the memory.

\begin{figure*}[tp]
\begin{tmplbox}[unbreakable]{Part of the trajectory behind the memory of Figure~\ref{fig:example-note}, labeled by segment}
\input{prompts/example_traj_flake8_display.tex}
\end{tmplbox}
\caption{A real trajectory split into segments (flake8-simplify, training split, frozen 27B actor).
Verbatim except \texttt{[...]} cuts and the omitted command fences; segment labels and step numbers
are ours.}
\label{fig:example-traj}
\end{figure*}

\section{Segment and Field Tests}\label{app:field-ablation}
App.~\ref{app:derivation} reads the result of every segment into a line of the memory
(Table~\ref{tab:derivation-channels}).
Rerun on the training half alone, the segment conditions support the same choices for the commands,
the repository content, the tests and the reasoning text, with Shapley values~\citep{shapley1953value,ghorbani2019datashapley} of $+3.09$, $+2.67$ and
$+2.40$ steps for the first three and a negative one for the reasoning text, and test commands with
their exit codes alone save $+4.03$ steps. The memory also lowers steps on two external sets
(App.~\ref{app:external}).
The step differences quoted in \S\ref{sec:attrib} are medians with BH-adjusted
$p^{*}{=}.0070$ for the commands ($56$/$27$) and $p^{*}{=}.0302$ for the repository content read
($52$/$34$).
Against the material it replaces, the memory
keeps the whole median step saving on linked boundaries, $2$ of $2$ steps, with $6\%$ of its bytes
($1{,}056$ against $17{,}320$ characters) and a ratio interval of $[0.50, 2.00]$, which meets a
retention margin of $0.50$, and gives up no measurable step ($47$/$39$, $p{=}.48$).

\input{appendix_drafts/app_derivation}

\section{Anchors}\label{app:anchors}
Definition~\ref{def:anchor} is implemented as a string matcher, the same one that marks memory-driven
steps in \S\ref{sec:cca}. A text is split into tokens, and a token is an anchor when it falls into
one of the four kinds of Table~\ref{tab:anchor-kinds}. Three filters keep generic strings out. Tokens
shorter than six characters are dropped, as are tokens whose every part is a common English or shell
word (\texttt{command-line}, \texttt{pytest}), and scaffolding found in nearly every repository
(\texttt{setup.py}, \texttt{conftest.py}, \texttt{reproduce.py} and paths under \texttt{/tmp}). A
\texttt{path:lines} reference is matched on its path, so the line range in a memory guides the next task
without being required for a match. For provenance (Eq.~\eqref{eq:prov}) the memory's anchors are further
reduced to those absent from the session so far and from what the memory-free sessions of the same
group reach in their first eight steps.

\begin{table}[H]
\centering
\scriptsize
\setlength{\tabcolsep}{3pt}
\caption{\textbf{The four kinds of anchor.} Examples are from the memory in Figure~\ref{fig:example-note}
except the node id, whose form is shown.}
\label{tab:anchor-kinds}
\begin{tabular}{@{}l l >{\raggedright\arraybackslash}p{2.45cm} >{\raggedright\arraybackslash}p{2.15cm}@{}}
\toprule
\textbf{Kind} & \textbf{Role} & \textbf{Rule} & \textbf{Example} \\
\midrule
File path & location & has a \texttt{/} and a letter, or a code extension & \texttt{flake8\_simplify/ rules/ast\_if.py} \\
Symbol & location & snake, camel, dotted or hyphenated, not all common words & \texttt{is\_main\_check}, \texttt{ast.Constant} \\
Test node id & check & contains \texttt{::}, its path and name also count & \texttt{tests/test\_x.py:: test\_y} \\
Command token & check & non-flag token inside a command or a \texttt{VERIFY}/\texttt{TESTS} field & \texttt{flake8}, \texttt{sim102} \\
\bottomrule
\end{tabular}
\end{table}

\begin{figure*}[tp]
\begin{tmplbox}[unbreakable]{A memory written in-session by the 27B actor (flake8-simplify, training split)}
\begin{Verbatim}
FILES: flake8_simplify/rules/ast_if.py
STATUS: submitted | TESTS: python -m pytest tests/ -q [exit 0]
VERIFY: python -m pytest tests/ -k sim102 -q [exit 0] | COVERS: flake8_simplify/rules/ast_if.py
VERIFY: python -m flake8 --select=SIM102 /tmp/test_sim102.py [exit 0] | COVERS: flake8_simplify/rules/ast_if.py
DID: added is_main_check() to skip SIM102 when outer if is `if __name__ == "__main__":` | WHERE: flake8_simplify/rules/ast_if.py:34-58 (get_sim102, is_main_check)
CONTEXT: rules live in flake8_simplify/rules/*.py, one file per AST node type; SIM102 is in ast_if.py | WHERE: flake8_simplify/rules/ast_if.py
CONTEXT: tests run via `python -m pytest tests/ -q`; full suite is 90 passed, 2 xfailed | WHERE: tests/
CONTEXT: to_source() in utils.py uses astor; constants AST_CONST_TYPES/STR_TYPES cover both ast.Constant and legacy ast.Str/ast.Num | WHERE: flake8_simplify/constants.py
\end{Verbatim}
\tcblower
\scriptsize\textbf{Anchors the matcher extracts (18).} \emph{Locations:}
\texttt{flake8\_simplify/rules/ast\_if.py}, \texttt{flake8\_simplify/rules}, \texttt{ast\_if.py},
\texttt{flake8\_simplify/constants.py}, \texttt{constants.py}, \texttt{utils.py}, \texttt{is\_main\_check},
\texttt{get\_sim102}, \texttt{AST\_CONST\_TYPES/STR\_TYPES}, \texttt{STR\_TYPES}, \texttt{ast.Constant},
\texttt{ast.Str/ast.Num}, \texttt{ast.Num}. \emph{Checks:} \texttt{flake8}, \texttt{sim102},
\texttt{SIM102}, \texttt{test\_sim102.py}, \texttt{tmp/test\_sim102.py}.
\end{tmplbox}
\caption{A real memory and its anchors. The memory has $889$ characters and $18$ anchors.}
\label{fig:example-note}
\end{figure*}

\section{Memory Templates}\label{app:templates}
Our memory is written by the acting model itself, in a final turn appended to the session that ends a
task, under the instruction in Figure~\ref{fig:note-instruction}, reproduced verbatim from the deployed
file (SHA-256 \texttt{02a36ea0\ldots}). A malformed memory gets one corrective retry, and a session that
still ends without a valid memory has it written in a separate call.

\begin{figure*}[tp]
\begin{tmplbox}[unbreakable]{Memory-Writing Instruction (final turn of the session)}
\begin{Verbatim}
Your work on this task is over; nothing you write now is executed. \textbf{Write a MEMORY NOTE} for
whoever works on this repository next. This is the last turn of the session.

\textbf{What the next session has, and what it does not:}
- It starts in the same working tree. Every change of this session is already committed, so it can
  read the diff itself (git show HEAD). Do NOT restate the diff.
- It runs every command in a fresh shell from the repository root: nothing under /tmp, no cd, and
  no environment variable survives from one command to the next.
- It may be working on a DIFFERENT task than this session. It cannot see this conversation.

The note has two blocks. Write the TASK block first and mark its scope explicitly.

\tmplhead{=== TASK BLOCK (applies only if the next task touches the files listed in FILES) ===}
\textbf{FILES:} \tmplph{<every file this session modified or ran tests for, comma-separated, verbatim paths>}
\textbf{STATUS:} \tmplph{<submitted | stopped at step limit | no code change>} | \textbf{TESTS:} \tmplph{<last full test command>} [last exit N]
\textbf{VERIFY:} \tmplph{<exact command as issued, runnable from the repository root in a fresh shell>} [exit N] | \textbf{COVERS:} \tmplph{<path>}[, \tmplph{<path>}]
\textbf{VERIFY:} \tmplph{<...>}            (one line per distinct command that checked something; keep the ones that FAILED too, with their nonzero exit code; most recent last)
\textbf{DID:} \tmplph{<what changed, one clause>} | \textbf{WHERE:} \tmplph{<path:lines (symbols)>}        (anchors only; never paste diff hunks)

\tmplhead{=== REPO BLOCK (facts that hold for any task in this repository) ===}
\textbf{CONTEXT:} \tmplph{<repository fact the next session would otherwise rediscover: how tests are run, layout, conventions, environment quirks>} | \textbf{WHERE:} \tmplph{<path:lines or the command that showed it>}

\textbf{Rules:}
- Copy paths, commands, symbol names and test ids VERBATIM from this session; never invent, round,
  or add a directory prefix that was never printed.
- A command that failed is worth recording exactly as much as one that passed: write its real exit
  code. Never round or invent an exit code.
- VERIFY commands must run unchanged from the repository root in a fresh shell: a repository test
  command (pytest path or test id) or a self-contained python -c. Never reference /tmp, a script
  that is not in the repository, or a prior cd. If the only check was a scratch script, write the
  repository test command that covers the change instead, or VERIFY: none | COVERS: \tmplph{<changed files>}.
- STATUS and DID describe THIS session's task only. They belong under FILES so that a session on
  unrelated files can skip them; never phrase them as facts about the repository.
- Do not write why you did things, do not summarise the conversation, do not quote test output.
  Test results are carried by exit codes, not by pasted text.
- Write CONTEXT lines only for facts this session actually established (a command that showed them,
  a file that states them). Never add a CONTEXT line to fill the block.
- \textbf{At most 10 lines and 1200 characters in total.} There is no fixed priority among the line types and
  no requirement to emit every type: spend the budget where THIS session's material actually
  warrants it.
- Output the note ONLY. No command, no code fence, no THOUGHT text, no explanation, and do not
  restate these instructions. This turn runs nothing.

\textbf{Line-start rule}, restated because it is the most common failure: every line of your output
must begin with one of exactly FILES:, STATUS:, VERIFY:, DID:, CONTEXT:. TESTS, WHERE and
COVERS are NEVER line starts -- they occur only as " | " continuations inside one of those
five lines.

\textbf{Shape of a finished note (placeholders, not values to copy):}

\textbf{FILES:} \tmplph{<path>}, \tmplph{<path>}
\textbf{STATUS:} \tmplph{<submitted|stopped at step limit|no code change>} | \textbf{TESTS:} \tmplph{<command>} [exit \tmplph{<N>}]
\textbf{VERIFY:} \tmplph{<command>} [exit \tmplph{<N>}] | \textbf{COVERS:} \tmplph{<path>}
\textbf{DID:} \tmplph{<one clause>} | \textbf{WHERE:} \tmplph{<path>}:\tmplph{<lines>} (\tmplph{<symbol>})
\textbf{CONTEXT:} \tmplph{<repository fact>} | \textbf{WHERE:} \tmplph{<path>}:\tmplph{<lines>}
\end{Verbatim}
\end{tmplbox}
\caption{The instruction under which the actor writes its memory, verbatim from the deployed file. It
states what the next session can see, fixes the memory's two blocks and five line types, sets the rules
and shows the finished shape. Bold, shading and gray placeholders are ours.}
\label{fig:note-instruction}
\end{figure*}

\section{Design Choices and Their Alternatives}\label{app:discussion}
This section sets out the alternatives to three design choices and what the paper's own measurements
say about them.

\paragraph{Adapting the actor rather than training a separate memory reader.}
A separate reader, or a retrieval policy, decides what memory to hand the actor, either before the task
or between its steps. In our setting the memory is short, at most ten lines, and already sits in the
actor's context, so the open question is not which memory to show but whether and when the actor acts
on it. Which steps use the memory emerges as the task unfolds. What a coding agent needs is often not
evident from the task description and surfaces as it explores and executes, so a single retrieval
keyed on the task input falls short of multi-step repository exploration~\citep{ircot2023,
yang2024sweagent, zhang2026sweexplore}, retrieval from the unfinished code alone misses what a draft
reveals~\citep{repocoder2023}, and memory retrieved per subtask during execution outperforms memory
retrieved by whole-task similarity~\citep{shen2026subtaskmem}. Under the detector that training uses,
credited steps concentrate late in a trajectory, where the actor runs the tests the memory names, while
early steps that match the memory are mostly steps the actor takes without any memory
(App.~\ref{app:prov-calib}). A reader that selects memory up front cannot see this, and a reader that
re-selects at every step would need the same per-step signal we give the actor. We therefore adapt the
actor itself and pay the measured value to the steps that use the memory. Retrieval stays compatible
with this choice, since a retrieval policy decides what memory enters the context and the adaptation
changes what the actor does with it.
Related controllers decide when to inject reminders (PMA; \citealp{pma2026}), whether to retrieve
or abstain (RSCB-MC; \citealp{rscbmc2026}), and how to navigate memory at different granularities
(NapMem; \citealp{napmem2026}). These choices concern memory access, whereas our update targets
execution after a short memory is already present.

\paragraph{A LoRA adapter rather than a separate model or full fine-tuning.}
A separate memory model leaves the actor untouched but cannot change how it acts on memory, and full
fine-tuning can move the actor's general coding behaviour. A LoRA adapter on frozen base weights is the
middle ground, a low-rank update that bounds how far the actor moves while letting it adapt its use of
memory. Whether general capability survives is a measurement, not a property of the adapter. On
SWE-bench Verified, with no memory in context, the adapted 27B and 9B actors resolve $4.5$ and $6.4$
points more than their untrained counterparts with $7\%$ and $22\%$ fewer steps
(Table~\ref{tab:ext-verified}), so on this benchmark the adaptation cost no general capability.

\paragraph{Paying the value through provenance rather than leaving it to the outcome reward.}
Outcome reward already favours whatever raises $R$, including a good use of memory, but it spreads
one scalar over every step of a session, and the steps that use the memory are about three percent of
them (App.~\ref{app:prov-calib}). Proposition~\ref{prop:route} shows that paid to every step the
measured value is a baseline, while paid through the gate it becomes a push on how often the actor acts
on the memory, sized and signed by the on-policy measurement. Figure~\ref{fig:modules} compares this
credit with whole-trajectory credit under the same reward. A gate read from how much the memory raised a
step's likelihood is the natural alternative to anchor provenance and carried no signal on this actor
(App.~\ref{app:prov-calib}).

\section{Related Work}\label{app:related}
\paragraph{Agent memory beyond coding.}
Agent memory encompasses reflective feedback and extracted experience~\citep{shinn2023reflexion,zhao2024expel},
reusable workflows and executable skills~\citep{wang2025awm,wang2024voyager}, and adaptive
inference-time experience~\citep{suzgun2025cheatsheet,tang2025agentkb}.
Other systems organize episodic or conversational information through reflection, retrieval, memory
hierarchies and linked records~\citep{park2023generative,zhong2024memorybank,packer2023memgpt,xu2025amem}.
MemOS and EverMemOS address memory organization and lifecycle management~\citep{memos2025,evermemos2026};
\citet{zhang2024memorysurvey} survey the broader design space.

\paragraph{Memory for coding agents.}
Memory systems keep different forms of experience: Mem0 extracts facts from conversations \citep{mem0_2025}, ReasoningBank and SWE-Exp distill lessons from past trajectories \citep{reasoningbank2025,sweexp2025}, Repository Memory and MemCoder mine commit history for localization and development knowledge \citep{repomem2026,memcoder2026}, and MemOp maintains a per-repository guidebook with a trained writer \citep{memop2026}.
These approaches emphasize retained experience or repository knowledge; our setting additionally
carries the actual edited repository state.
AdaRepair-Mem selects repair experience by coverage, quality and repair stage~\citep{adarepairmem2026}.
Repository context files also require empirical evaluation: AGENTS.md studies find that added
instructions need not improve success and can increase cost~\citep{agentsmd2026}.
Persistent project information appears in deployed assistants through Claude Code's auto memory and
Codex's cross-session memory generation and injection~\citep{claudecode_memory_docs2026,codex_memory_config2026}.
Within one task, SWE-MeM and CompactionRL learn context compression jointly with issue resolution \citep{swemem2026,li2026compactionrl}, and the concurrent MemTrace re-validates stored execution traces against the evolving repository \citep{memtrace2026}; these memories do not outlive the task.
CAMG-RL likewise learns file-based memory operations across context boundaries within an
episode~\citep{luo2026camg}.
Between the two lies the boundary a deployed agent meets, a memory written before the next task is known and read in the repository state the last task left.

\paragraph{Broader memory evaluation.}
LoCoMo and LongMemEval test long-term conversational memory~\citep{maharana2024locomo,wu2025longmemeval},
StateMemBench tests evolving-state tracking~\citep{statemembench2026}, and MemoryArena couples memory
with interdependent multi-session actions~\citep{memoryarena2026}.
LongCodeBench evaluates long-context code comprehension and repair~\citep{longcodebench2025}.
These settings probe memory retention and use, while our evaluation measures the effort saved on an
unseen coding task from inherited repository state.

\paragraph{Evaluating memory on executable tasks.}
SWE-bench grades each issue from its own snapshot \citep{swebench2024}, whereas ChainSWE chains fixes in a shared workspace \citep{chainswe2026} and SWE-Milestone reports resolve falling from above 80\% to 38\% once state is carried \citep{swemilestone2026}; SWEContextBench and Handoff Debt study reuse across related issues and the cost of resuming an interrupted one \citep{swecontextbench2026,handoffdebt2026}.
Memory-specific evaluations isolate the retained information: MemOp reruns the source task with and without a candidate memory \citep{memop2026}, VibeMemBench pairs memory-off and memory-on runs of one target whose history is summarised with the target in view \citep{vibemembench2026}, and Decision-Aware Memory Cards score evidence by a heuristic estimate of its effect on the next action \citep{memcards2026}.
Related proxies include interventions on candidate memories for answer quality~\citep{cmi2026},
answer-conditioned information gain~\citep{infomem2026}, and Shapley-based context filtering for code
completion~\citep{reposhapley2026}. MemDelta emphasizes controlled model and retrieval comparisons
and write-path cost accounting~\citep{memdelta2026}.
These evaluations complement our paired runs of an unseen next task from the inherited tree,
which measure resolution and effort and attribute savings to trajectory segments (\S\ref{sec:replay}).

\paragraph{Learning memory construction and use.}
Reinforcement learning has been applied to both sides of the memory interface.
Writers are trained from outcome reward or neighbourhood-level marginal utility \citep{memalpha2025,memoryr1_2025,ye2026umem}, with finer credit from rerolling at a shared memory state \citep{memoryr2_2026}, hindsight-filtered utility on writing tokens \citep{yan2026himpo}, or token-level attribution \citep{attrimem2026}; RoMeRL notes the outcome-broadcast trap but treats paired rollouts as infeasible \citep{romerl2026}.
On the reading side, ECHO routes outcome credit to the turns a policy selects back into context \citep{xie2026echo}, and MemCalib localises credit by atom ablation \citep{memcalib2026}.
End-to-end memory learning also includes recurrent consolidation in MEM1~\citep{zhou2025mem1},
chunk-wise processing in MemAgent~\citep{yu2025memagent}, self-memory policies in MemPO~\citep{li2026mempo},
and operation-tree credit in Mem-T~\citep{yue2026memt}.
These methods optimize different memory interfaces and credit granularities. Our contribution pairs
on-policy valuation on executable next tasks with provenance-gated routing of the additional
memory-value term, while retaining outcome credit at every step (\S\ref{sec:grpo}).

\paragraph{Credit assignment beyond memory.}
Classical RL assigns hindsight credit to past actions~\citep{harutyunyan2019hca} or uses
counterfactual baselines to separate action contributions~\citep{foerster2018coma,mesnard2021cca}.
For language models, process supervision supplies intermediate labels~\citep{lightman2024verify},
VinePPO estimates step values with sampled continuations~\citep{kazemnejad2025vineppo},
GiGPO groups repeated states for local advantages~\citep{feng2025gigpo}, and C3 replays downstream
continuations after message interventions in agent teams~\citep{c3credit2026}.
These methods address action or message utility; our provenance gate identifies traceable memory
use and routes measured memory value without estimating each step's causal effect.

\paragraph{Efficient coding and policy training.}
SWE-Gym and SWE-smith expand executable training tasks and agent trajectories~\citep{pan2025swegym,yang2025swesmith},
while SWE-RL trains on software-evolution data~\citep{wei2025swerl}.
SWE-Edit reduces overhead through editing-interface decomposition~\citep{sweedit2026};
L1 and efficient-reasoning RL shape generation length and inference compute~\citep{aggarwal2025l1,arora2025efficient}.
These approaches complement our focus on information reuse and memory use across task boundaries.

\end{document}

%% file: src/math_commands.tex
\usepackage{amsmath,amsfonts,bm}

\def\eqref#1{equation~\ref{#1}}

\def\1{\bm{1}}

\DeclareMathAlphabet{\mathsfit}{\encodingdefault}{\sfdefault}{m}{sl}
\SetMathAlphabet{\mathsfit}{bold}{\encodingdefault}{\sfdefault}{bx}{n}



%% file: tables/table1_main.tex
\begin{table*}[!t]
\centering
\scriptsize
\setlength{\tabcolsep}{0pt}\renewcommand{\arraystretch}{0.84}
\caption{\textbf{Cost and outcome of each memory layer for the open-weight actor models on the test set.} Resolve rate
and steps on linked and unlinked boundaries (App.~\ref{app:setup}) and overall, cache-aware tokens per session of the
actor model and of the memory module (Writer), and resolve per thousand actor or total tokens. Tags give the change
against the no-memory row of the same actor model. Gray values in parentheses are medians over sessions. \coremem{} w/o training is our memory on the frozen actor model and
\coremem{} the post-trained actor model with it; replay counts and tests are in App.~\ref{app:setup}, with the definition of every column.}
\label{tab:ledger}
\resizebox{\textwidth}{!}{%
\begin{tabular}{l@{\hspace{8pt}}c@{\hspace{8pt}}c@{\hspace{10pt}}c@{\hspace{8pt}}c@{\hspace{10pt}}c@{\hspace{18pt}}c@{\hspace{28pt}}c@{\hspace{10pt}}c@{\hspace{8pt}}c@{\hspace{18pt}}c@{\hspace{10pt}}c@{\hspace{18pt}}}
\toprule
\multirow{2}{*}{\textbf{Memory layer}} & \multicolumn{2}{c}{\textbf{Linked}} & \multicolumn{2}{c}{\textbf{Unlinked}}
& \multicolumn{2}{c}{\textbf{Overall}} & \multicolumn{3}{c}{\textbf{Tokens per session (K)}}
& \multicolumn{2}{c}{\textbf{Res.\ per K tokens}} \\
& Res.\ (\%) $\uparrow$ & Steps $\downarrow$ & Res.\ (\%) $\uparrow$ & Steps $\downarrow$ & Res.\ (\%) $\uparrow$
& Steps $\downarrow$ & Actor $\downarrow$ & Writer & Total $\downarrow$ & Actor $\uparrow$ & Total $\uparrow$ \\
\midrule
\multicolumn{12}{c}{\cellcolor{gray!15}\textbf{Qwen3.5-9B}} \\
\midrule
No memory & 9.7 & 98.4\,{\color{gray}\tiny(91.1)} & 19.4 & 98.1\,{\color{gray}\tiny(91.2)} & 14.0 & 98.2\,{\color{gray}\tiny(91.2)} & 98.3\,{\color{gray}\tiny(98.9)} & -- & 98.3\,{\color{gray}\tiny(98.9)} & 0.152 & 0.152 \\
Full history & 11.4 & 70.0\,{\color{gray}\tiny(60.5)} & 17.9 & 76.4\,{\color{gray}\tiny(68.5)} & 14.2\hang{\dg{+2\%}} & 72.8\,{\color{gray}\tiny(65.1)}\hang{\dg{\textminus26\%}} & 85.9\,{\color{gray}\tiny(79.9)} & -- & 85.9\,{\color{gray}\tiny(79.9)}\hang{\dg{\textminus13\%}} & 0.156 & 0.156\hang{\dg{+3\%}} \\
\cmidrule{1-12}
Mem0 & 12.2 & 94.3\,{\color{gray}\tiny(86.0)} & 21.8 & 93.1\,{\color{gray}\tiny(85.2)} & 16.4\hang{\dg{+17\%}} & 93.8\,{\color{gray}\tiny(86.0)}\hang{\dg{\textminus5\%}} & 96.5\,{\color{gray}\tiny(94.8)} & 14.9\,{\color{gray}\tiny(14.4)} & 111.4\,{\color{gray}\tiny(109.1)}\hang{\dl{+13\%}} & 0.175 & 0.151\hang{\dz{0\%}} \\
MemOp & 11.9 & 93.5\,{\color{gray}\tiny(85.8)} & 19.0 & 96.8\,{\color{gray}\tiny(86.3)} & 15.0\hang{\dg{+7\%}} & 94.9\,{\color{gray}\tiny(86.1)}\hang{\dg{\textminus3\%}} & 99.3\,{\color{gray}\tiny(96.4)} & 32.4\,{\color{gray}\tiny(33.1)} & 131.7\,{\color{gray}\tiny(131.4)}\hang{\dl{+34\%}} & 0.147 & 0.111\hang{\dl{\textminus27\%}} \\
ReasoningBank & 12.0 & 92.3\,{\color{gray}\tiny(83.6)} & 17.9 & 96.7\,{\color{gray}\tiny(90.3)} & 14.6\hang{\dg{+4\%}} & 94.3\,{\color{gray}\tiny(85.2)}\hang{\dg{\textminus4\%}} & 102.3\,{\color{gray}\tiny(94.0)} & 63.7\,{\color{gray}\tiny(65.8)} & 166.1\,{\color{gray}\tiny(163.1)}\hang{\dl{+69\%}} & 0.143 & 0.088\hang{\dl{\textminus42\%}} \\
\cmidrule{1-12}
\textbf{\coremem{} w/o training} & 12.5 & 95.6\,{\color{gray}\tiny(87.2)} & 19.2 & 94.6\,{\color{gray}\tiny(88.6)} & 15.5\hang{\dg{+11\%}} & 95.1\,{\color{gray}\tiny(87.2)}\hang{\dg{\textminus3\%}} & 101.7\,{\color{gray}\tiny(104.6)} & 10.0\,{\color{gray}\tiny(7.5)} & 111.7\,{\color{gray}\tiny(116.5)}\hang{\dl{+14\%}} & 0.140 & 0.128\hang{\dl{\textminus16\%}} \\
\textbf{\coremem{}} & 14.2 & 70.2\,{\color{gray}\tiny(64.2)} & 22.2 & 75.4\,{\color{gray}\tiny(69.3)} & 17.7\hang{\dg{+27\%}} & 72.4\,{\color{gray}\tiny(65.1)}\hang{\dg{\textminus26\%}} & 75.0\,{\color{gray}\tiny(65.4)} & 5.0\,{\color{gray}\tiny(2.0)} & 79.9\,{\color{gray}\tiny(70.8)}\hang{\dg{\textminus19\%}} & 0.236 & 0.222\hang{\dg{+46\%}} \\
\midrule
\multicolumn{12}{c}{\cellcolor{gray!15}\textbf{Qwen3.8-27B-FP8}} \\
\midrule
No memory & 19.1 & 30.9\,{\color{gray}\tiny(22.8)} & 23.4 & 40.5\,{\color{gray}\tiny(30.4)} & 21.0 & 35.1\,{\color{gray}\tiny(26.0)} & 32.2\,{\color{gray}\tiny(24.7)} & -- & 32.2\,{\color{gray}\tiny(24.7)} & 0.653 & 0.653 \\
Full history & 20.7 & 27.0\,{\color{gray}\tiny(21.3)} & 24.6 & 39.8\,{\color{gray}\tiny(33.5)} & 22.4\hang{\dg{+7\%}} & 32.6\,{\color{gray}\tiny(23.8)}\hang{\dg{\textminus7\%}} & 39.0\,{\color{gray}\tiny(29.5)} & -- & 39.0\,{\color{gray}\tiny(29.5)}\hang{\dl{+21\%}} & 0.574 & 0.574\hang{\dl{\textminus12\%}} \\
\cmidrule{1-12}
Mem0 & 20.1 & 32.5\,{\color{gray}\tiny(26.6)} & 23.4 & 40.3\,{\color{gray}\tiny(30.0)} & 21.5\hang{\dg{+2\%}} & 35.9\,{\color{gray}\tiny(28.2)}\hang{\dl{+2\%}} & 34.2\,{\color{gray}\tiny(25.4)} & 34.5\,{\color{gray}\tiny(28.4)} & 68.7\,{\color{gray}\tiny(53.4)}\hang{\dl{+113\%}} & 0.632 & 0.314\hang{\dl{\textminus52\%}} \\
MemOp & 22.5 & 29.4\,{\color{gray}\tiny(22.2)} & 24.6 & 38.9\,{\color{gray}\tiny(36.8)} & 23.4\hang{\dg{+12\%}} & 33.5\,{\color{gray}\tiny(24.8)}\hang{\dg{\textminus4\%}} & 32.3\,{\color{gray}\tiny(23.8)} & 18.2\,{\color{gray}\tiny(14.7)} & 50.4\,{\color{gray}\tiny(38.0)}\hang{\dl{+57\%}} & 0.727 & 0.465\hang{\dl{\textminus29\%}} \\
ReasoningBank & 23.8 & 30.1\,{\color{gray}\tiny(24.6)} & 27.0 & 39.8\,{\color{gray}\tiny(31.8)} & 25.2\hang{\dg{+20\%}} & 34.3\,{\color{gray}\tiny(27.7)}\hang{\dg{\textminus2\%}} & 32.5\,{\color{gray}\tiny(26.2)} & 35.0\,{\color{gray}\tiny(28.7)} & 67.5\,{\color{gray}\tiny(54.7)}\hang{\dl{+110\%}} & 0.778 & 0.375\hang{\dl{\textminus43\%}} \\
\cmidrule{1-12}
\textbf{\coremem{} w/o training} & 22.5 & 26.9\,{\color{gray}\tiny(21.2)} & 24.5 & 38.0\,{\color{gray}\tiny(32.9)} & 23.4\hang{\dg{+11\%}} & 31.8\,{\color{gray}\tiny(22.4)}\hang{\dg{\textminus10\%}} & 30.4\,{\color{gray}\tiny(23.0)} & 2.0\,{\color{gray}\tiny(1.7)} & 32.4\,{\color{gray}\tiny(24.7)}\hang{\dz{0\%}} & 0.772 & 0.724\hang{\dg{+11\%}} \\
\textbf{\coremem{}} & 25.1 & 26.2\,{\color{gray}\tiny(18.0)} & 26.5 & 33.4\,{\color{gray}\tiny(25.0)} & 25.7\hang{\dg{+22\%}} & 29.2\,{\color{gray}\tiny(20.0)}\hang{\dg{\textminus17\%}} & 26.6\,{\color{gray}\tiny(18.5)} & 1.9\,{\color{gray}\tiny(1.7)} & 28.5\,{\color{gray}\tiny(20.3)}\hang{\dg{\textminus11\%}} & 0.966 & 0.902\hang{\dg{+38\%}} \\
\midrule
\multicolumn{12}{c}{\cellcolor{gray!15}\textbf{Gemma-4-31B}} \\
\midrule
No memory & 4.3 & 30.4\,{\color{gray}\tiny(23.8)} & 20.6 & 29.3\,{\color{gray}\tiny(21.7)} & 11.5 & 29.9\,{\color{gray}\tiny(23.2)} & 32.4\,{\color{gray}\tiny(23.9)} & -- & 32.4\,{\color{gray}\tiny(23.9)} & 0.353 & 0.353 \\
Full history & 7.4 & 25.7\,{\color{gray}\tiny(18.5)} & 17.5 & 29.1\,{\color{gray}\tiny(19.8)} & 11.8\hang{\dg{+3\%}} & 27.2\,{\color{gray}\tiny(19.2)}\hang{\dg{\textminus9\%}} & 31.8\,{\color{gray}\tiny(25.8)} & -- & 31.8\,{\color{gray}\tiny(25.8)}\hang{\dg{\textminus2\%}} & 0.371 & 0.371\hang{\dg{+5\%}} \\
\cmidrule{1-12}
Mem0 & 4.9 & 28.7\,{\color{gray}\tiny(24.3)} & 18.3 & 27.4\,{\color{gray}\tiny(21.7)} & 10.8\hang{\dl{\textminus6\%}} & 28.1\,{\color{gray}\tiny(23.0)}\hang{\dg{\textminus6\%}} & 34.0\,{\color{gray}\tiny(26.4)} & 13.2\,{\color{gray}\tiny(13.3)} & 47.2\,{\color{gray}\tiny(39.2)}\hang{\dl{+46\%}} & 0.322 & 0.230\hang{\dl{\textminus35\%}} \\
MemOp & 4.9 & 26.8\,{\color{gray}\tiny(20.0)} & 23.0 & 31.3\,{\color{gray}\tiny(22.5)} & 12.8\hang{\dg{+12\%}} & 28.8\,{\color{gray}\tiny(20.5)}\hang{\dg{\textminus4\%}} & 34.6\,{\color{gray}\tiny(26.2)} & 14.9\,{\color{gray}\tiny(13.5)} & 49.6\,{\color{gray}\tiny(39.6)}\hang{\dl{+53\%}} & 0.371 & 0.259\hang{\dl{\textminus27\%}} \\
ReasoningBank & 7.4 & 30.6\,{\color{gray}\tiny(23.3)} & 20.6 & 30.3\,{\color{gray}\tiny(22.2)} & 13.2\hang{\dg{+15\%}} & 30.4\,{\color{gray}\tiny(23.0)}\hang{\dl{+2\%}} & 33.0\,{\color{gray}\tiny(26.0)} & 32.3\,{\color{gray}\tiny(28.9)} & 65.3\,{\color{gray}\tiny(54.9)}\hang{\dl{+102\%}} & 0.399 & 0.202\hang{\dl{\textminus43\%}} \\
\cmidrule{1-12}
\textbf{\coremem{} w/o training} & 6.8 & 28.2\,{\color{gray}\tiny(21.3)} & 19.8 & 28.5\,{\color{gray}\tiny(23.3)} & 12.5\hang{\dg{+9\%}} & 28.3\,{\color{gray}\tiny(22.0)}\hang{\dg{\textminus5\%}} & 31.0\,{\color{gray}\tiny(23.8)} & 2.9\,{\color{gray}\tiny(1.3)} & 33.9\,{\color{gray}\tiny(26.3)}\hang{\dl{+5\%}} & 0.405 & 0.371\hang{\dg{+5\%}} \\
\textbf{\coremem{}} & 8.6 & 28.0\,{\color{gray}\tiny(21.8)} & 21.4 & 29.1\,{\color{gray}\tiny(22.0)} & 14.2\hang{\dg{+24\%}} & 28.5\,{\color{gray}\tiny(21.8)}\hang{\dg{\textminus5\%}} & 27.2\,{\color{gray}\tiny(21.7)} & 2.2\,{\color{gray}\tiny(1.3)} & 29.3\,{\color{gray}\tiny(23.9)}\hang{\dg{\textminus9\%}} & 0.527 & 0.486\hang{\dg{+38\%}} \\
\bottomrule
\end{tabular}}
\end{table*}

%% file: tables/table_anchor_use.tex
\begin{table}[!t]
\centering
\small
\setlength{\tabcolsep}{3pt}
\newcommand{\vci}[3]{#1\,{\tiny[#2, #3]}}
\caption{\textbf{Anchors the next task can act on, per memory layer.} 27B frozen actor model, test
set; an anchor is \emph{used} when it occurs only in the memory and appears in a message of the next
task's session; brackets give $95\%$ intervals over the test chains.}
\label{tab:anchors}
\begin{tabular}{l|ccc}
\toprule
\textbf{Memory layer} & {\scriptsize\shortstack{\textbf{Anchors}\\\textbf{per 1K chars}}}
& {\scriptsize\shortstack{\textbf{Used anchors}\\\textbf{per memory}}}
& {\scriptsize\shortstack{\textbf{Used anchors}\\\textbf{per 1K chars}}} \\
\midrule
Mem0          & \vci{9.7}{9.4}{10.1}   & \vci{0.15}{0.09}{0.21} & \vci{0.14}{0.08}{0.21} \\
MemOp         & \vci{14.4}{13.6}{15.2} & \vci{2.06}{1.59}{2.51} & \vci{0.92}{0.72}{1.11} \\
ReasoningBank & \vci{8.5}{7.8}{9.1}    & \vci{0.65}{0.51}{0.80} & \vci{0.19}{0.14}{0.24} \\
\midrule
\begin{tabular}[c]{@{}l@{}}\textbf{\coremem{}}\\\textbf{w/o training}\end{tabular} & \vci{\textbf{15.8}}{15.0}{16.7} & \vci{\textbf{2.24}}{1.67}{2.87} & \vci{\textbf{1.32}}{1.00}{1.68} \\
\bottomrule
\end{tabular}
\end{table}

%% file: appendix_drafts/app_train_dynamics.tex

\begin{figure*}[t]
\centering
\includegraphics[width=\textwidth]{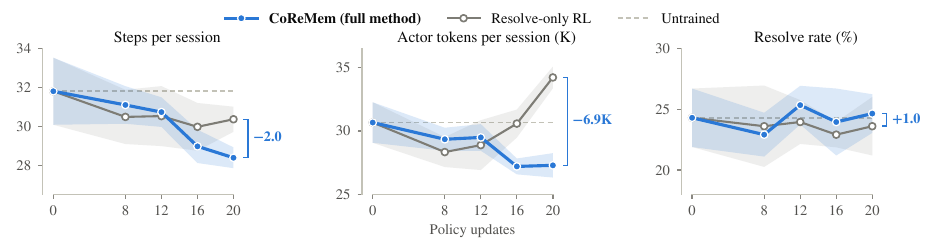}
\caption{Training dynamics of the 27B actor on the test sessions with the memory. Lines are means over
three replays and bands their standard deviation. Update $0$ is the untrained actor, also drawn as the
dashed line. The number beside each panel is the paired difference at update $20$.}
\label{fig:train-dynamics}
\end{figure*}

\paragraph{Training dynamics.} The training reward cannot compare the two recipes, because the full method
adds a step term and a value credit that resolve-only RL does not have. We therefore follow both runs on held-out
sessions. Figure~\ref{fig:train-dynamics} reads the 27B actor after $8$, $12$, $16$ and $20$ updates of
the full method and of resolve-only RL, which is the M1$+$M2 recipe without its step term, on the
$96$ test sessions with the memory, with three replays per checkpoint, all in one window. Both runs sum
their loss over tokens and share learning rate, update count, group shape and memory pool. From update
$12$ on, the full method keeps lowering its effort, by $0.25$ steps per
update, while resolve-only RL stays near $30$ steps and its actor tokens rise back above those of the
untrained actor. At update $20$, the checkpoint of the main tables, the full method uses $2.0$ fewer steps
($95\%$ CI $[-4.1, +0.4]$) and $6.9$K fewer actor tokens ($[-11.2, -3.1]$) than resolve-only RL, and
resolves $1.0$ points more ($[-2.4, +4.5]$).

%% file: tables/table_closed.tex
\begin{table*}[!t]
\centering
\scriptsize
\setlength{\tabcolsep}{0pt}\renewcommand{\arraystretch}{0.84}
\caption{\textbf{Cost and outcome of each memory layer for the API-served models on the test set.} Overall columns
as in Table~\ref{tab:ledger}, with the list-price cost per session and resolve per dollar.}
\label{tab:closed}
\resizebox{\textwidth}{!}{%
\begin{tabular}{l@{\hspace{8pt}}c@{\hspace{22pt}}c@{\hspace{30pt}}c@{\hspace{10pt}}c@{\hspace{8pt}}c@{\hspace{22pt}}c@{\hspace{10pt}}c@{\hspace{22pt}}c@{\hspace{22pt}}c@{\hspace{22pt}}}
\toprule
\multirow{2}{*}{\textbf{Memory layer}} & \multicolumn{2}{c}{\textbf{Outcome}} & \multicolumn{3}{c}{\textbf{Tokens per session (K)}}
& \multicolumn{2}{c}{\textbf{Res.\ per K tokens}} & \multicolumn{2}{c}{\textbf{List price}} \\
& Res.\ (\%) $\uparrow$ & Steps $\downarrow$ & Actor $\downarrow$ & Writer & Total $\downarrow$ & Actor $\uparrow$
& Total $\uparrow$ & USD $\downarrow$ & Res.\ per USD $\uparrow$ \\
\midrule
\multicolumn{10}{c}{\cellcolor{gray!15}\textbf{Claude Opus 5.5}} \\
\midrule
No memory & 38.5 & 12.6\,{\color{gray}\tiny(12.0)} & 21.5\,{\color{gray}\tiny(15.7)} & -- & 21.5\,{\color{gray}\tiny(15.7)} & 1.791 & 1.791 & 0.2370 & 162 \\
Full history & 40.6\hang{\dg{+5\%}} & 10.4\,{\color{gray}\tiny(9.0)}\hang{\dg{\textminus17\%}} & 27.0\,{\color{gray}\tiny(23.5)} & -- & 27.0\,{\color{gray}\tiny(23.5)}\hang{\dl{+25\%}} & 1.505 & 1.505\hang{\dl{\textminus16\%}} & 0.2384\hang{\dl{+1\%}} & 170\hang{\dg{+5\%}} \\
\cmidrule{1-10}
Mem0 & 39.6\hang{\dg{+3\%}} & 13.0\,{\color{gray}\tiny(12.0)}\hang{\dl{+3\%}} & 18.7\,{\color{gray}\tiny(16.2)} & 16.6\,{\color{gray}\tiny(17.8)} & 35.4\,{\color{gray}\tiny(34.8)}\hang{\dl{+64\%}} & 2.112 & 1.119\hang{\dl{\textminus37\%}} & 0.2893\hang{\dl{+22\%}} & 137\hang{\dl{\textminus16\%}} \\
MemOp & 39.6\hang{\dg{+3\%}} & 17.8\,{\color{gray}\tiny(12.5)}\hang{\dl{+41\%}} & 21.7\,{\color{gray}\tiny(16.0)} & 14.4\,{\color{gray}\tiny(12.1)} & 36.1\,{\color{gray}\tiny(28.2)}\hang{\dl{+68\%}} & 1.821 & 1.096\hang{\dl{\textminus39\%}} & 0.3630\hang{\dl{+53\%}} & 109\hang{\dl{\textminus33\%}} \\
ReasoningBank & 40.6\hang{\dg{+5\%}} & 14.2\,{\color{gray}\tiny(12.0)}\hang{\dl{+13\%}} & 20.9\,{\color{gray}\tiny(16.8)} & 27.9\,{\color{gray}\tiny(24.6)} & 48.8\,{\color{gray}\tiny(41.3)}\hang{\dl{+127\%}} & 1.939 & 0.832\hang{\dl{\textminus54\%}} & 0.4028\hang{\dl{+70\%}} & 101\hang{\dl{\textminus38\%}} \\
\cmidrule{1-10}
\textbf{\coremem{} w/o training} & 38.5\hang{\dz{0\%}} & 11.8\,{\color{gray}\tiny(10.0)}\hang{\dg{\textminus6\%}} & 18.1\,{\color{gray}\tiny(13.7)} & 2.0\,{\color{gray}\tiny(2.0)} & 20.1\,{\color{gray}\tiny(15.7)}\hang{\dg{\textminus7\%}} & 2.134 & 1.919\hang{\dg{+7\%}} & 0.2048\hang{\dg{\textminus14\%}} & 188\hang{\dg{+16\%}} \\
\midrule
\multicolumn{10}{c}{\cellcolor{gray!15}\textbf{GPT-5.6 Luna}} \\
\midrule
No memory & 23.6 & 17.9\,{\color{gray}\tiny(16.7)} & 51.7\,{\color{gray}\tiny(43.7)} & -- & 51.7\,{\color{gray}\tiny(43.7)} & 0.461 & 0.461 & 0.0176 & 1349 \\
Full history & 24.0\hang{\dg{+1\%}} & 18.6\,{\color{gray}\tiny(15.3)}\hang{\dl{+4\%}} & 79.8\,{\color{gray}\tiny(66.7)} & -- & 79.8\,{\color{gray}\tiny(66.7)}\hang{\dl{+54\%}} & 0.306 & 0.306\hang{\dl{\textminus34\%}} & 0.0263\hang{\dl{+50\%}} & 918\hang{\dl{\textminus32\%}} \\
\cmidrule{1-10}
Mem0 & 22.6\hang{\dl{\textminus4\%}} & 17.5\,{\color{gray}\tiny(15.3)}\hang{\dg{\textminus2\%}} & 49.8\,{\color{gray}\tiny(40.4)} & 13.0\,{\color{gray}\tiny(12.9)} & 62.7\,{\color{gray}\tiny(53.0)}\hang{\dl{+21\%}} & 0.456 & 0.361\hang{\dl{\textminus22\%}} & 0.0199\hang{\dl{+13\%}} & 1134\hang{\dl{\textminus16\%}} \\
MemOp & 22.9\hang{\dl{\textminus3\%}} & 17.5\,{\color{gray}\tiny(15.8)}\hang{\dg{\textminus2\%}} & 49.7\,{\color{gray}\tiny(38.3)} & 14.0\,{\color{gray}\tiny(12.6)} & 63.8\,{\color{gray}\tiny(49.7)}\hang{\dl{+23\%}} & 0.470 & 0.363\hang{\dl{\textminus21\%}} & 0.0204\hang{\dl{+16\%}} & 1121\hang{\dl{\textminus17\%}} \\
ReasoningBank & 23.3\hang{\dl{\textminus1\%}} & 17.9\,{\color{gray}\tiny(15.5)}\hang{\dz{0\%}} & 55.6\,{\color{gray}\tiny(44.8)} & 29.6\,{\color{gray}\tiny(26.2)} & 85.3\,{\color{gray}\tiny(72.0)}\hang{\dl{+65\%}} & 0.419 & 0.273\hang{\dl{\textminus41\%}} & 0.0258\hang{\dl{+47\%}} & 902\hang{\dl{\textminus33\%}} \\
\cmidrule{1-10}
\textbf{\coremem{} w/o training} & 25.7\hang{\dg{+9\%}} & 16.3\,{\color{gray}\tiny(14.7)}\hang{\dg{\textminus9\%}} & 48.9\,{\color{gray}\tiny(39.6)} & 3.6\,{\color{gray}\tiny(2.5)} & 52.5\,{\color{gray}\tiny(44.2)}\hang{\dl{+2\%}} & 0.526 & 0.491\hang{\dg{+7\%}} & 0.0181\hang{\dl{+3\%}} & 1421\hang{\dg{+5\%}} \\
\midrule
\multicolumn{10}{c}{\cellcolor{gray!15}\textbf{DeepSeek-V4-Flash}} \\
\midrule
No memory & 21.9 & 90.0\,{\color{gray}\tiny(87.3)} & 84.4\,{\color{gray}\tiny(70.6)} & -- & 84.4\,{\color{gray}\tiny(70.6)} & 0.262 & 0.262 & 0.0363 & 605 \\
Full history & 22.4\hang{\dg{+2\%}} & 76.6\,{\color{gray}\tiny(72.4)}\hang{\dg{\textminus15\%}} & 75.8\,{\color{gray}\tiny(66.0)} & -- & 75.8\,{\color{gray}\tiny(66.0)}\hang{\dg{\textminus10\%}} & 0.297 & 0.297\hang{\dg{+14\%}} & 0.0319\hang{\dg{\textminus12\%}} & 705\hang{\dg{+16\%}} \\
\cmidrule{1-10}
Mem0 & 21.9\hang{\dz{0\%}} & 94.2\,{\color{gray}\tiny(91.9)}\hang{\dl{+5\%}} & 87.5\,{\color{gray}\tiny(76.2)} & 12.5\,{\color{gray}\tiny(12.5)} & 100.0\,{\color{gray}\tiny(88.1)}\hang{\dl{+18\%}} & 0.252 & 0.220\hang{\dl{\textminus16\%}} & 0.0393\hang{\dl{+8\%}} & 558\hang{\dl{\textminus8\%}} \\
MemOp & 21.4\hang{\dl{\textminus2\%}} & 87.5\,{\color{gray}\tiny(85.6)}\hang{\dg{\textminus3\%}} & 81.7\,{\color{gray}\tiny(68.0)} & 28.8\,{\color{gray}\tiny(30.4)} & 110.5\,{\color{gray}\tiny(99.4)}\hang{\dl{+31\%}} & 0.263 & 0.194\hang{\dl{\textminus26\%}} & 0.0396\hang{\dl{+9\%}} & 542\hang{\dl{\textminus10\%}} \\
ReasoningBank & 23.1\hang{\dg{+6\%}} & 97.2\,{\color{gray}\tiny(96.2)}\hang{\dl{+8\%}} & 91.8\,{\color{gray}\tiny(78.4)} & 58.9\,{\color{gray}\tiny(62.5)} & 150.7\,{\color{gray}\tiny(145.3)}\hang{\dl{+78\%}} & 0.256 & 0.155\hang{\dl{\textminus41\%}} & 0.0468\hang{\dl{+29\%}} & 497\hang{\dl{\textminus18\%}} \\
\cmidrule{1-10}
\textbf{\coremem{} w/o training} & 23.6\hang{\dg{+8\%}} & 78.2\,{\color{gray}\tiny(77.1)}\hang{\dg{\textminus13\%}} & 75.9\,{\color{gray}\tiny(57.6)} & 6.4\,{\color{gray}\tiny(1.5)} & 82.3\,{\color{gray}\tiny(61.6)}\hang{\dg{\textminus3\%}} & 0.313 & 0.289\hang{\dg{+10\%}} & 0.0330\hang{\dg{\textminus9\%}} & 717\hang{\dg{+18\%}} \\
\bottomrule
\end{tabular}}
\end{table*}

%% file: tables/table_verified.tex
\begin{table*}[t]
\centering
\small
\setlength{\tabcolsep}{5pt}
\caption{\textbf{The post-trained actor on SWE-bench Verified.} All $500$ instances, one task each, no
memory, $3$ replays per row. The untrained row serves the zero-update adapter on the same path, and tags
give the change against it ($^{**}$ for a paired $99\%$ interval that excludes zero). Raw counts tokens
with no credit for caching, and Limit hit is the share of instances that end at the step limit.}
\label{tab:ext-verified}
\resizebox{\textwidth}{!}{%
\begin{tabular}{l|l|ccccc|c}
\toprule
\textbf{Actor} & \textbf{Checkpoint} & \textbf{Resolve (\%)} $\uparrow$ & \textbf{Steps mean (med.)} $\downarrow$ & \textbf{Actor tok (K) mean (med.)} $\downarrow$ & \textbf{Raw tok (K)} $\downarrow$ & \textbf{Limit hit (\%)} $\downarrow$ & \textbf{Res/Tok (actor)} $\uparrow$ \\
\midrule
\multirow{2}{*}{\shortstack[l]{Qwen3.8-27B-FP8\\\citep{qwen38_2026}}} & untrained & 69.2\sdsub{1.9} & \mmc{28.3\sdsub{0.2}}{21.2} & \mmc{24.7\sdsub{0.6}}{17.4} & 383.0\sdsub{15.2} & 7.0\sdsub{0.4} & 2.80\sdsub{0.10} \\
 & \textbf{post-trained} & 73.7\sdsub{0.6}\hang{\dg{+7\%**}} & \mmc{26.2\sdsub{0.3}\hang{\dg{\textminus7\%**}}}{20.5} & \mmc{22.5\sdsub{0.2}\hang{\dg{\textminus9\%**}}}{17.2} & 313.7\sdsub{1.5}\hang{\dg{\textminus18\%**}} & 2.5\sdsub{0.1}\hang{\dg{\textminus64\%**}} & 3.27\sdsub{0.05}\hang{\dg{+17\%**}} \\
\midrule
\multirow{2}{*}{\shortstack[l]{Qwen3.5-9B\\\citep{qwen35_2026}}} & untrained & 49.1\sdsub{2.2} & \mmc{90.1\sdsub{1.3}}{86.2} & \mmc{93.5\sdsub{0.4}}{86.3} & 2471.9\sdsub{20.6} & 7.1\sdsub{1.9} & 0.53\sdsub{0.03} \\
 & \textbf{post-trained} & 55.5\sdsub{0.3}\hang{\dg{+13\%**}} & \mmc{69.9\sdsub{1.4}\hang{\dg{\textminus22\%**}}}{63.8} & \mmc{72.9\sdsub{1.3}\hang{\dg{\textminus22\%**}}}{64.1} & 1586.1\sdsub{70.4}\hang{\dg{\textminus36\%**}} & 0.1\sdsub{0.1}\hang{\dg{\textminus99\%**}} & 0.76\sdsub{0.02}\hang{\dg{+43\%**}} \\
\bottomrule
\end{tabular}}

\end{table*}

%% file: tables/placeholder_external_memory.tex
\label{tab:ext-memory}

\begin{table*}[!t]
\centering
\scriptsize
\setlength{\tabcolsep}{0pt}\renewcommand{\arraystretch}{0.84}
\caption{\textbf{The memory layer off the training distribution: successive tasks of one repository on an external dataset.} SWEContextBench (SCB) pairs are two-task chains of one repository in one image, the next task opening at an independent checkout from the first, so only memory crosses the border. 
\coremem{} w/o training writes its memory under the instruction of Figure~\ref{fig:note-instruction}; we updated the writer instructions by
adding that the next session starts from a new checkout without the earlier changes, that the memory should not include a summary of or reason for the changes, and that the memory should only include one clause per \texttt{DID} line. The adapted instructions were written after observing the first
runs on the set, whose memories were getting cut off because of note length.}
\label{tab:scb-ledger}
\resizebox{\textwidth}{!}{%
\begin{tabular}{l@{\hspace{8pt}}c@{\hspace{8pt}}c@{\hspace{10pt}}c@{\hspace{8pt}}c@{\hspace{10pt}}c@{\hspace{18pt}}c@{\hspace{28pt}}c@{\hspace{10pt}}c@{\hspace{8pt}}c@{\hspace{18pt}}c@{\hspace{10pt}}c@{\hspace{18pt}}}
\toprule
\multirow{2}{*}{\textbf{Memory layer}} & \multicolumn{2}{c}{\textbf{Linked}} & \multicolumn{2}{c}{\textbf{Unlinked}}
& \multicolumn{2}{c}{\textbf{Overall}} & \multicolumn{3}{c}{\textbf{Tokens per session (K)}}
& \multicolumn{2}{c}{\textbf{Res.\ per K tokens}} \\
& Res.\ (\%) $\uparrow$ & Steps $\downarrow$ & Res.\ (\%) $\uparrow$ & Steps $\downarrow$ & Res.\ (\%) $\uparrow$
& Steps $\downarrow$ & Actor $\downarrow$ & Writer & Total $\downarrow$ & Actor $\uparrow$ & Total $\uparrow$ \\
\midrule
\multicolumn{12}{c}{\cellcolor{gray!15}\textbf{Qwen3.8-27B-FP8, agent without network access} {\scriptsize ($n{=}4$ replays)}} \\
\midrule
No memory & 31.6 & 59.2\,{\color{gray}\tiny(61.2)} & 19.6 & 44.6\,{\color{gray}\tiny(37.5)} & 28.1 & 54.9\,{\color{gray}\tiny(54.4)} & 89.4\,{\color{gray}\tiny(60.8)} & -- & 89.4\,{\color{gray}\tiny(60.8)} & 0.340 & 0.340 \\
No memory, post-trained actor & 33.8 & 51.2\,{\color{gray}\tiny(48.0)} & 21.4 & 39.9\,{\color{gray}\tiny(34.4)} & 30.2\hang{\dg{+7\%}} & 47.9\,{\color{gray}\tiny(44.8)}\hang{\dg{\textminus13\%}} & 76.2\,{\color{gray}\tiny(47.8)} & -- & 76.2\,{\color{gray}\tiny(47.8)}\hang{\dg{\textminus15\%}} & 0.406 & 0.406\hang{\dg{+19\%}} \\
\cmidrule{1-12}
\textbf{\coremem{} w/o training} & 36.8 & 40.3\,{\color{gray}\tiny(36.4)} & 19.6 & 43.2\,{\color{gray}\tiny(39.1)} & 31.8\hang{\dg{+13\%}} & 41.2\,{\color{gray}\tiny(39.1)}\hang{\dg{\textminus25\%}} & 61.5\,{\color{gray}\tiny(41.4)} & 5.2\,{\color{gray}\tiny(2.0)} & 66.7\,{\color{gray}\tiny(43.4)}\hang{\dg{\textminus25\%}} & 0.561 & 0.517\hang{\dg{+52\%}} \\
\textbf{\coremem{}} & 40.4 & 35.7\,{\color{gray}\tiny(32.0)} & 16.1 & 34.8\,{\color{gray}\tiny(29.9)} & 33.3\hang{\dg{+19\%}} & 35.4\,{\color{gray}\tiny(30.4)}\hang{\dg{\textminus35\%}} & 57.6\,{\color{gray}\tiny(35.1)} & 4.0\,{\color{gray}\tiny(1.9)} & 61.6\,{\color{gray}\tiny(37.2)}\hang{\dg{\textminus31\%}} & 0.628 & 0.583\hang{\dg{+71\%}} \\
\bottomrule
\end{tabular}}
\end{table*}

%% file: tables/table_openhands.tex
\begin{table}[t]
\centering
\footnotesize
\setlength{\tabcolsep}{3.5pt}
\caption{\textbf{Memory layers under the OpenHands harness.} The 27B actor model on $94$ of the $96$ next
tasks of the test split, mean of two runs, frozen in the upper rows and post-trained in the last row, where it
is served through its LoRA adapter. Tokens are raw counts per session with no credit for prefix caching, memory
module included, and do not compare with Table~\ref{tab:ledger}.}
\label{tab:openhands}
\begin{tabular}{@{}lcccc@{}}
\toprule
Memory layer & \shortstack{Resolve\\(\%)} & Steps & \shortstack{Tokens\\(K)} & \shortstack{Resolve\\per M\\tokens} \\
\midrule
No memory & 10.6 & 33.8 & 897 & 11.9 \\
Full history & 14.9 & 41.8 & 1704 & 8.8 \\
Mem0 & 11.2 & 38.3 & 1176 & 9.5 \\
MemOp & 14.4 & 38.5 & 1273 & 11.9 \\
ReasoningBank & 21.8 & 48.0 & 1943 & 11.3 \\
\coremem{} w/o training & 16.5 & 38.6 & 1240 & 13.3 \\
\midrule
\coremem{} & 21.3 & 45.9 & 1647 & 13.0 \\
\bottomrule
\end{tabular}
\end{table}

%% file: prompts/example_traj_flake8_display.tex
\begin{Verbatim}[fontsize=\scriptsize,breaklines=true,breaksymbolleft={},commandchars=\^\{\}]
^stepmark{step 3}
^chlab{chdropbg}{REASONING TEXT:}   THOUGHT: It's a directory. Let me look at the rules directory.
^chlab{chkeybg}{COMMAND:} cd /flake8-simplify && ls flake8_simplify/rules/ && grep -rn "SIM102" flake8_simplify/   [exit 0]
^chlab{chkeybg}{REPOSITORY CONTENT:}
  __init__.py
  ast_assign.py
  [...]
  ast_if.py
  [...]
  flake8_simplify/rules/ast_if.py:12:    SIM102 = "SIM102 Use a single if-statement instead of nested if-statements"
  flake8_simplify/rules/ast_if.py:34:    errors.append((node.lineno, node.col_offset, SIM102))

^ellipsisline{[... steps 4-6 read ast_if.py, constants.py and utils.py ...]}

^stepmark{step 7}
^chlab{chdropbg}{REASONING TEXT:}   THOUGHT: Now I'll add a helper function and use it in get_sim102. Let me edit the file.
^chlab{chkeybg}{COMMAND:} cd /flake8-simplify && python - <<'EOF'   [exit 0]
         path = "flake8_simplify/rules/ast_if.py"
         [...]
         def is_main_check(node: ast.If) -> bool:
             """Check if the if-statement is ``if __name__ == "__main__":``."""
         [...]
         EOF
^chlab{chredbg}{TEST OUTPUT (the script contains an assert):}
  done

^stepmark{step 8}
^chlab{chdropbg}{REASONING TEXT:}   THOUGHT: Now let me verify the fix works with a test script.
^chlab{chkeybg}{COMMAND:} cd /flake8-simplify && cat > /tmp/test_sim102.py <<'EOF'   [exit 0]
         if __name__ == "__main__":
             if foo():
                 pass
         EOF
         python -m flake8 --select=SIM102 /tmp/test_sim102.py; echo "exit: $?"
         cat > /tmp/test_sim102b.py <<'EOF'
         if a:
             if b:
                 pass
         EOF
         python -m flake8 --select=SIM102 /tmp/test_sim102b.py; echo "exit: $?"
^chlab{chkeybg}{REPOSITORY CONTENT (flake8 is not a test runner):}
  exit: 0
  /tmp/test_sim102b.py:1:1: SIM102 Use a single if-statement instead of nested if-statements
  exit: 1

^stepmark{step 9}
^chlab{chdropbg}{REASONING TEXT:}   THOUGHT: The fix works. The `__main__` case no longer triggers, while normal nested ifs
         still do. Let me run the existing test suite for SIM102 to ensure nothing breaks.
^chlab{chkeybg}{COMMAND:} cd /flake8-simplify && python -m pytest tests/ -k "sim102" -q 2>&1 | tail -20   [exit 0]
^chlab{chredbg}{TEST OUTPUT:}
  [...]
  4 passed, 88 deselected in 0.37s

^stepmark{step 10}
^chlab{chdropbg}{REASONING TEXT:}   THOUGHT: All SIM102 tests pass. Let me run the full test suite to be safe.
^chlab{chkeybg}{COMMAND:} cd /flake8-simplify && python -m pytest tests/ -q 2>&1 | tail -5   [exit 0]
^chlab{chredbg}{TEST OUTPUT:}
  [...]
  90 passed, 2 xfailed in 0.48s

^stepmark{step 11}
^chlab{chdropbg}{REASONING TEXT:}   THOUGHT: All tests pass. Let me create the patch.
^chlab{chkeybg}{COMMAND:} cd /flake8-simplify && git diff -- flake8_simplify/rules/ast_if.py > patch.txt && cat patch.txt   [exit 0]
^chlab{chredbg}{FINAL DIFF:}
  --- a/flake8_simplify/rules/ast_if.py
  +++ b/flake8_simplify/rules/ast_if.py
  @@ -31,10 +31,33 @@ def get_sim102(node: ast.If) -> List[Tuple[int, int, str]]:
       if not is_pattern_1:
           return errors
  +    if is_main_check(node):
  +        return errors
  [...]

^chlab{chredbg}{OUTCOME:} Submitted, 12 steps
^chlab{chdropbg}{SCRATCH FILES:} /tmp/test_sim102.py, /tmp/test_sim102b.py (gone at the next task); patch.txt (kept)
\end{Verbatim}

%% file: appendix_drafts/app_derivation.tex

\section{From Raw History to the Memory}\label{app:derivation}
This appendix records how each line of the \coremem{} memory was obtained from the raw session, and where
each decision came from. The derivation ran in five steps. The first three are mechanical, the fourth
arranges the fields into the memory's two blocks, and the fifth validates the result. Table~\ref{tab:derivation-channels}
maps every segment measurement to the line it produced.

\paragraph{Step 1, segments.} The current task's session is split by the role each piece of text plays,
namely the agent's reasoning (C1), the commands it issued (C2), the repository content it read (C3), the
feedback of its tests (C4), the change it left (C5), its outcome (C6) and its scratch files (C7). The
outcome segment is cut in two, the outcome the writer can observe and the verdict of the hidden tests,
and the latter may never enter a memory. Three compressed forms are added as probes, the failing commands
with their output, the diff's file list, and the test commands with their exit codes but no output.

\paragraph{Step 2, values.} Each segment subset is valued by replaying the next task from the fixed
state the current task left (\S\ref{sec:apcv}). The subsets are no segment, all segments twice, all but
one segment and one segment alone, plus the three probes. A segment's
\emph{carried share} is the fraction of the whole material's median step saving that it keeps when
injected alone. The readings below are on linked boundaries (App.~\ref{app:setup}) unless stated
otherwise.

\paragraph{Step 3, inventory and constraints.} A decision program turns the values into the
list of fields and the constraints on how each is written. It emitted \texttt{VERIFY}, \texttt{CONTEXT},
\texttt{DID}, \texttt{TESTS} and a one-line reason field, and fixed four constraints.
The location
of a change is written as anchors with its diff body banned, a failed check stays a \texttt{VERIFY} entry
with its nonzero exit code, a test result is the exit code of its command, and the hidden-test verdict
never appears.

\paragraph{Step 4, layout.} The fields go into two blocks. The task block opens with \texttt{FILES}, the
files the session modified or tested, and holds \texttt{STATUS}, \texttt{VERIFY} and \texttt{DID}, which describe
this session's task; the repository block holds the \texttt{CONTEXT} facts that hold for any task. The
reasoning segment carried no step value, so the reason field was dropped, and \texttt{TESTS} was merged into
the \texttt{STATUS} line.

\paragraph{Step 5, validation.} The memory is compared with the raw material it replaces and keeps its
whole median step saving with $6\%$ of its bytes (App.~\ref{app:field-ablation}).

\begin{table*}[t]
\centering\small
\setlength{\tabcolsep}{4pt}
\begin{tabular}{@{}>{\raggedright\arraybackslash}p{0.13\textwidth}>{\raggedright\arraybackslash}p{0.33\textwidth}>{\raggedright\arraybackslash}p{0.22\textwidth}>{\raggedright\arraybackslash}p{0.26\textwidth}@{}}
\toprule
\textbf{Segment} & \textbf{Measurement on linked boundaries} & \textbf{Decision} & \textbf{Line of the memory} \\
\midrule
C1 reasoning & Deleting it costs no step (34 helped, 45 hurt) and saves 1.14K tokens ($p{=}.023$) & Step 3 kept a reason field; Step 4 dropped it & None; the instruction forbids reasons and summaries \\
C2 commands & Deleting it costs a median 2 steps (56/27, BH $p^{*}{=}.0070$); alone it carries the whole saving; the failing commands alone carry half & Kept; failed checks stay in it with their exit code & \texttt{VERIFY} exact command [exit $N$] \texttt{|} \texttt{COVERS} paths \\
C3 repository content & Deleting it costs a median step (52/34, BH $p^{*}{=}.0302$) and saves 0.63K tokens & Kept as located facts & \texttt{CONTEXT} fact \texttt{|} \texttt{WHERE} path:lines, in the repository block \\
C4 test feedback & Test commands with exit codes beat the full test output by a median step (53/30, BH $p^{*}{=}.045$) & A test result is the exit code of its command & \texttt{TESTS} last full test command [last exit $N$]; no pasted output \\
C5 change & Deleting the diff body costs no step (47/44) and saves 0.94K tokens ($p{=}.021$); the file list matches the full diff ($p{=}.097$) & Anchors only, body banned & \texttt{DID} one clause \texttt{|} \texttt{WHERE} path:lines (symbols); the change is committed and is not restated \\
C6 outcome & No effect overall (45/33); helps when the current task resolved (24/11) & Observable part kept, hidden-test verdict banned; Step 4 places it in the task block & \texttt{STATUS} submitted, stopped or no change, inside the task block \\
C7 scratch & 38\% of the paths a session writes sit under \texttt{/tmp} and are gone by the next command; the manifest carries no step value (36/48) & No line; a validity rule on \texttt{VERIFY} & \texttt{VERIFY} runs unchanged from the repository root, with no \texttt{/tmp} and no \texttt{cd} \\
\bottomrule
\end{tabular}
\caption{\textbf{From segments to lines.} Results of the segment study
(App.~\ref{app:field-ablation}). Counts are boundaries where the segment helped and hurt the
next task. BH marks the multiplicity correction across segments.}
\label{tab:derivation-channels}
\end{table*}

\paragraph{Scope of the procedure.} Steps 2 and 5 use only paired replay and the next task's cost, so
they apply wherever sessions follow one another on persistent state. Steps 1 and 3 need a split of the
session by role and a definition of an anchor, which on ChainSWE are the seven segments and the
locations and checks of Definition~\ref{def:anchor}.